\documentclass{article}

\PassOptionsToPackage{numbers, compress}{natbib}

\usepackage[preprint]{neurips_2023}

\usepackage{enumitem}

\usepackage[utf8]{inputenc} 
\usepackage[T1]{fontenc}    
\usepackage{hyperref}       
\usepackage{url}            
\usepackage{booktabs}       
\usepackage{amsfonts}       
\usepackage{nicefrac}       
\usepackage{microtype}      
\usepackage{xcolor}         
\usepackage{amssymb}
\usepackage{pifont}
\usepackage{graphicx}
\usepackage{amsmath}
\usepackage{caption}
\usepackage{graphicx}
\usepackage{booktabs}
\usepackage{floatrow}
\usepackage{subcaption}
\usepackage{wrapfig}
\usepackage{caption}
\newfloatcommand{capbtabbox}{table}[][\FBwidth]
\newcommand{\ryan}[1]{{{\color{red} cite}}}
\newcommand{\myparagraph}[1]{\noindent\textbf{#1.}}

\title{Variational Information Pursuit with Large Language and Multimodal Models for Interpretable Predictions}
                  
\author{Kwan Ho Ryan Chan\\
Johns Hopkins University \\
\texttt{kchan49@jhu.edu}
\And
Aditya Chattopadhyay\\
Johns Hopkins University \\
\texttt{achatto1@jhu.edu}
\And
\And
Benjamin D.~Haeffele\\
Johns Hopkins University\\
\texttt{bhaeffele@jhu.edu}
\And
Ren\'e Vidal\\
University of Pennsylvania\\ 
\texttt{vidalr@seas.upenn.edu}
}

\begin{document}

\maketitle

\begin{abstract}

Variational Information Pursuit (V-IP) is a framework for making interpretable predictions by design by sequentially selecting a short chain of task-relevant, user-defined and interpretable queries about the data that are most informative for the task. While using queries related with semantic concepts allows for built-in interpretability in predictive models, applying V-IP to any task requires data samples with concept-labeling by domain experts, limiting the application of V-IP to small-scale tasks where manual data annotation is feasible. In this work, we extend the V-IP framework with Foundational Models (FMs) to address this limitation. More specifically, we use a two-step process, by first leveraging Large Language Models (LLMs) to generate a sufficiently large candidate set of task-relevant interpretable concepts, then using multimodal models to annotate each data sample by semantic similarity with each concept in the generated concept set. While other interpretable-by-design frameworks such as Concept Bottleneck Models (CBMs) require an additional step of removing repetitive and non-discriminative concepts to have good interpretability and test performance, we mathematically and empirically justify that, with a sufficiently informative and task-relevant query (concept) set, the proposed FM+V-IP method does not require any type of concept filtering. In addition, we show that FM+V-IP with LLM generated concepts can achieve better test performance than V-IP with human annotated concepts, demonstrating the effectiveness of LLMs at generating efficient query sets. Finally, when compared to other interpretable-by-design frameworks such as CBMs, FM+V-IP can achieve competitive test performance using fewer number of concepts/queries in both cases with filtered or unfiltered concept sets.
\end{abstract}

\section{Introduction}\label{section: Introduction}

Interpretable-by-design methods incorporate interpretability by requiring model predictions to be solely dependent on (intermediate) representations that correspond to semantic concepts. The set of concepts and their annotations in data is often created and labeled by domain experts for the task. As a result, the prediction is made based on the composition of interpretable concepts that explains the model's underlying decision making. 
One line of research on interpretable-by-design frameworks is Variational Information Pursuit (V-IP):
The user first defines a set of queries, which are concept-related functions of the data such as ``Is the car red?''. The model then makes a prediction by selecting a small number of informative queries that maximize the mutual information between the task and query-answers. Another line of work is Concept Bottleneck Models (CBMs)~\citep{koh2020concept}, for which inputs are first nonlinearly mapped to an interpretable concept space, then linearly mapped to the target's label space
via sparse-promoting regularizers such as the $\ell_1$-norm so that each prediction becomes interpretable by only using a small number of concepts. An example of both frameworks is shown Figure~\ref{fig:vip_vs_cbm}. Although interpretable-by-design seems ideal, built-in interpretability is not cost-free: CBMs and V-IP have been traditionally limited to small- to medium- scale datasets. Applying interpretable-by-design frameworks to large-scale datasets such as ImageNet entails overcoming three main challenges: 


\begin{table}[t]
\caption{Comparisons between this work and other interpretable-by-design frameworks.}
\resizebox{\textwidth}{!}{\begin{tabular}{l|cccc}
\toprule
Methods & \begin{tabular}[c]{@{}c@{}}Annotates samples\\ automatically\end{tabular} & \begin{tabular}[c]{@{}c@{}}Do not require\\ concept filtering\end{tabular} & \begin{tabular}[c]{@{}c@{}}Operates on large\\ query sets\end{tabular} & \begin{tabular}[c]{@{}c@{}}Predict with variable\\ no. of queries/concepts\end{tabular} \\
\midrule
CBM & No & No & No & No \\
Lf-CBM & \textbf{Yes} & \textbf{Yes} & \textbf{Yes} & No \\
LaBo & \textbf{Yes} & No & \textbf{Yes} & No \\
V-IP & No & No & No & \textbf{Yes} \\
\midrule
FM+V-IP (this work) & \textbf{Yes} & \textbf{Yes} & \textbf{Yes} & \textbf{Yes} \\
\bottomrule
\end{tabular}
}

\label{table:highlight}
\end{table}

\textbf{Challenge 1: Generating a set of interpretable queries.} Interpretable-by-design frameworks require the user to first define a set of interpretable concepts that are sufficiently informative for the task. While good interpretable concepts are often created by domain experts for optimal interpretability, 
when the number of classes increases or when the task is too complex, manually generating interpretable and task-specific concept sets becomes challenging. To address this issue, we propose to use Large Language Models (LLMs) such as GPT-3~\citep{brown2020language} as the machine expert in place of human domain experts for extending to tasks of any scale. Since LLMs are trained on a large amount of text, we operate on the premise that LLMs are machine experts that have learned discriminative information about the data and labels for any given task, 
thus they can produce large query sets from fairly simple prompts.

\textbf{Challenge 2: Annotating every sample per concept.} Once a sufficiently informative, task-relevant, interpretable concept set is generated (either by LLMs or by humans), the concepts needs to be related to the data for interpretable predictions. For instance, in CUB-200~\citep{wah2011caltech}, each of the expert-defined 312 concepts is individually labelled for every one of the 11,788 images. However, in the case of a large-scale dataset, annotating every sample by hand for datasets with up to millions of samples is infeasible without any machine assistance. In this work, we propose to use CLIP, a Vision-Language Pre-trained model (VLPs) trained on a large number of image-text pairs, to annotate every sample in an efficient manner.
CLIP computes a similarity score for each pair of image-text input, where a high score indicates the presence of a certain concept and vice versa.
By taking advantage of VLPs, we allow faster data annotations that enables interpretable-by-design frameworks for large-scale data. 

\textbf{Challenge 3: The need for concept filtering.} With a suitable query set generated by LLMs and per-sample annotated data from VLPs, theoretically one should be able to scale interpretable-by-design frameworks to data and tasks of any scale. Nonetheless, recent works such as Label-free CBMs (Lf-CBM)~\citep{oikarinen2023label} show that filtering out concepts from the initial concept set (also known as candidate set) is needed in order to achieve good predictive performance and interpretability. Concepts that are too similar to other concepts semantically, for example, should be removed since one of them represents redundant information. Nonetheless, there is no principled way to judge what ad-hoc concept filtering methods should or should not be used in the final concept set. In this work, we argue that the formulation of V-IP can perform interpretable predictions \textit{without} the need to perform any concept filtering: Computing the mutual information for the task and next most informative query naturally disregards uninformative queries that would have been filtered out.

\myparagraph{Paper contributions} The main contributions of this work are the following: 
\begin{enumerate}[leftmargin=*]
\item We extend V-IP to large query sets (300 to 45K concepts) and 
large visual datasets (50K to 1M images) by using Foundational Models to generate and answer queries. We also empirically show that LLMs and VLPs are capable of generating informative query sets for interpretable predictions.
\item We demonstrate that V-IP can perform interpretable predictions without the need of ad-hoc concept filtering methods when compared to other interpretable-by-design frameworks such as Lf-CBM. 
\item We further demonstrate that V-IP with query set of size up to 45K  queries/concepts can achieve competitive test performance with a smaller number of queries/concepts than other methods such as Lf-CBM and LaBo~\citep{yang2022language}. The major differences between the different frameworks are summarized in Table~\ref{table:highlight}. 

\end{enumerate}



\section{Related Work}\label{section: Related Work}
\myparagraph{Explanations without language} 
Post-hoc explanation methods including feature attribution methods, such as LIME~\citep{ribeiro2016should} and SHAP~\citep{lundberg2017unified}, as well as gradient-based methods, GradCAM~\citep{selvaraju2017grad, chattopadhay2018grad} and Integrated Gradients~\citep{sundararajan2017axiomatic} generate explanations directly on feature maps/pixels that do not involve natural language. Similarly for interpretable-by-design methods, image classification experiments in sequential selection methods such as V-IP~\citep{chattopadhyay2023variational,covert2023learning} use all possible $8 \times 8$ image patches as their query set. Overall, using input data features as ``concepts'' versus natural language concepts poses the debate of what semantics the selected features represent. Conversely, in this work, we focus solely on natural language concepts, providing a more direct and colloquial description of the underlying models' decisions.

\myparagraph{Foundational Models} OpenAI's GPT-3~\citep{brown2020language} is a Large Language Model (LLM) trained on large amounts of unlabeled text in a self-supervised manner. Recent benchmarks have demonstrated that LLMs can perform complex human tasks, such as passing the Bar Exam~\citep{bommarito2022gpt}, showing signs that LLMs are capable of learning complex concepts and general understanding~\citep{bubeck2023sparks}. Adding other data modalities, CLIP~\citep{radford2021learning} is a Vision-Language Pre-trained models (VLPs) trained on a large dataset of image-text pairs. CLIP consists of an image encoder and a text encoder that output corresponding image and text embeddings whose dot-products encode cross-modal similarities between images and texts. Furthermore, improvements on CLIP achieve state-of-the-art performance in tasks such as zero-shot image classification~\citep{menon2022visual}, and related works have found use-cases of CLIP as an architecture backbone for human-level tasks such as Visual Question Answering~\citep{eslami2021does, parelli2023clip} and Video Question Answering~\citep{ye2023video}. Also known as Foundational Models (FMs), LLMs and VLPs have demonstrate strong capabilities to learn complex concepts, which further motivates the use of LLMs to generate interpretable concepts and the use VLPs to relate similarity between representations of data across different modalities. 


\myparagraph{Concept Bottleneck Models} 
The interpretable-by-design framework (and its extensions) that we mainly compare with in this work is Concept Bottleneck Models (CBMs)~\cite{koh2020concept}. A CBM mainly consist of two parts: 1) A nonlinear function $f: x \mapsto c$ that maps data to a concept feature space, where each feature corresponds to the relevance of the input sample and a concept from a pre-defined set of interpretable concepts; and 2) a linear function $g: c \mapsto y$ that maps from concept scores to task labels. $f$ is often a deep network such as ResNet-18~\citep{he2016deep}, while the predictor $g$ is chosen to be a simple model such as a linear network or a decision tree. Similar to V-IP, while every prediction is made interpretable, applications of CBMs are also limited to small- to medium-scale tasks where data with dense concept annotations are available. To address this, extensions of CBMs such as Post-hoc CBM~\citep{yuksekgonul2022post}, Lf-CBM~\citep{oikarinen2023label} and Language in a Bottle (LaBo)~\citep{yang2022language} use LLMs such as GPT-3 or WordNet~\citep{miller1998wordnet} to generate concept sets, and leverage neuron-labeling methods, such as MILAN~\citep{hernandez2022natural} and CLIP-Dissect~\citep{oikarinen2022clip}, or VLPs such as CLIP~\citep{radford2021learning} to annotate visual data with texts.

\section{Methods}\label{section: Methods}
Our method can mainly be divided into three main parts: 1) query set generation with GPT-3; 2) answering queries using CLIP; and 3) performing interpretable predictions with Variational Information Pursuit (V-IP)~\citep{chattopadhyay2023variational}. We begin by first introducing the background on the V-IP framework, including its original the generative approach to doing IP~\citep{chattopadhyay2022interpretable}, the sufficiency criteria, and IP algorithm. 

\subsection{Background: Information Pursuit and Variational Information Pursuit}

\myparagraph{IP} The Information Pursuit framework was first introduced in \citet{chattopadhyay2022interpretable} as an interpretable-by-design framework to perform interpretable predictions for any given task. Let $X: \Omega \rightarrow \mathcal{X}$ and $Y: \Omega \rightarrow \mathcal{Y}$ denote the random variables for input data and corresponding labels/outputs, and $\Omega$ be the underlying sample space were all random variables are defined. The user first define a set $Q$ of task-specific and interpretable queries $q: \mathcal{X} \rightarrow \mathcal{A}$, where $q(x) \in \mathcal{A}$ is the answer to the query $q \in Q$ evaluated at $x \in \mathcal{X}$. For all data-label pairs $(x, y) \in \mathcal{X} \times \mathcal{Y}$, We say that the query set $Q$ is sufficient for $Y$ when
\begin{equation}
    P(y \mid x) = P(y \mid \{x' \in \mathcal{X} : q(x') = q(x)\}).
\end{equation}
An implication of this is that a query set $Q$ is insufficient if we cannot estimate the posterior $P(y \mid x)$ properly given all query answers of $x$. With a sufficient query set $Q$, the IP algorithm is described as follows: Given a data point $x^\text{obs}$, the algorithm selects a sequence of most informative queries, until all remaining queries are nearly uninformative:
\begin{equation}
\label{alg:IP}
\begin{split}
    q_1 &= \text{IP}(\emptyset) = \mathrm{argmax}_{q \in Q}{I(q(X) ; Y)}; \\
    q_{k+1} &= \text{IP}(\{q_i, q_i(x^{\text{obs}})\}_{1:k}) = \mathrm{argmax}_{q \in Q} I(q(X); Y \mid q_{1:k}(x^{\textrm{obs}})). 
    \end{split}
\end{equation} 
 Here $q_{k+1}$ denotes the query selected at step $k+1$, given history $q_{1:k}(x^{\text{obs}})$. $I$ denotes mutual information. The IP algorithm terminates at the the stopping criteria, defined as the posterior exceeding a threshold $P( Y \mid q_{1:k}(x^{\text{obs}}) > 1 - \epsilon$ with a certain pre-determined $\epsilon$. The first formulation proposed in \citet{chattopadhyay2022interpretable} requires learning a generative model for $P(Q(X), Y)$ to estimate the mutual information terms. However, the method is not scalable to large-scale tasks due to the large computational cost during inference. To alleviate this issue, the authors later proposed V-IP, where the posterior distribution is learned directly using discriminative models, as described next.


\begin{wrapfigure}[12]{l}{0.47\textwidth}
\centering
    \includegraphics[width=\textwidth]{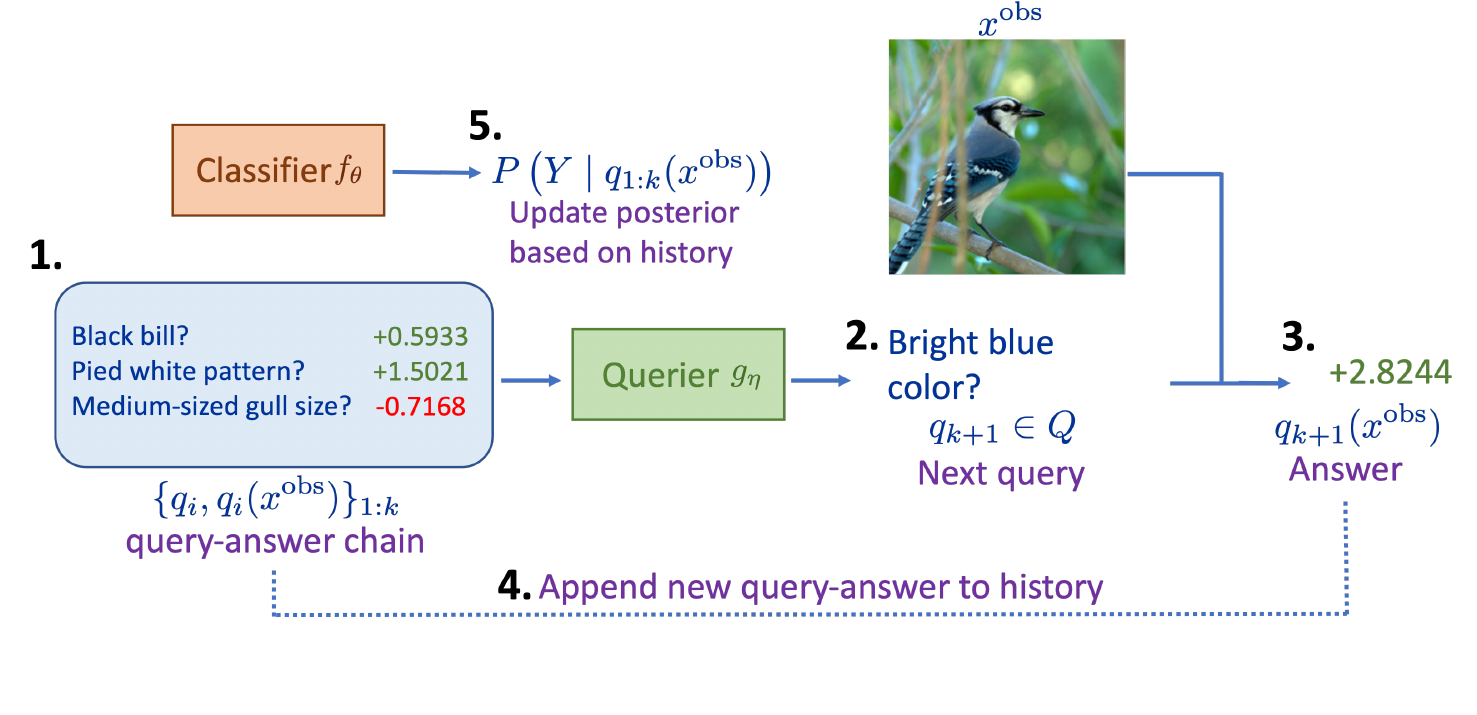}
    \caption{Overview of V-IP.}
    \label{fig:vip-diagram}
\end{wrapfigure}
\myparagraph{V-IP} Variational Information Pursuit (V-IP) was later introduced as a variational approach to IP~\citep{chattopadhyay2023variational}. V-IP
defines a predictor $f: S \rightarrow Y$ and a querier $g: S \rightarrow Q$ that map a query-answer chain $S$ of any length to the posterior distribution of $Y$ and the most informative next query given $S$, respectively. Parameterized by deep networks $\theta$ and $\eta$, $f_\theta$ and $g_\eta$ are trained by sampling random query-answer chains and optimizing the following V-IP objective: 
\begin{align*}
    \min_{\theta, \eta} \quad &\mathbb{E}_{X, S}[D_{KL}( P(Y | X) \| P_\theta (Y | q_\eta(X), S)] \label{eq:v-ip} \\
    \text{where} \quad & q_\eta := g_\eta(S) \notag\\
    &P(Y \mid q_{\eta}(X), S) := f_\theta(\{q_\eta, q_\eta(X) \cup S\}) \notag
\end{align*}
Importantly, \citet{chattopadhyay2023variational} shows that selecting the query $q_{k+1}$ with the optimal querier $g_\eta^*$ given any history $S$ is exactly equivalent to selecting the query with the maximum mutual information in Equation~\eqref{alg:IP}. Hence, performing inference using V-IP can be done with the trained querier $g_\eta^*$ in-place of computing the mutual information at each query:
\begin{equation}
\label{eq: V-IP algorithm}
\begin{split}
    q_1 &= g_\eta^*(\emptyset) = \mathrm{argmax}_{q \in Q}{I(q(X) ; Y)}; \\
    q_{k+1} &= g_\eta^*(\{q_i, q_i(x^{\text{obs}})\}_{1:k}) = \mathrm{argmax}_{q \in Q} I(q(X); Y \mid q_{1:k}(x^{\textrm{obs}})). 
    \end{split}
\end{equation} 
Figure~\ref{fig:vip-diagram} shows a flowchart of performing inference with V-IP. Empirical results show that V-IP is up to 100x faster in computational speed than the previous generative approach, while being able to scale up to larger scale datasets such as CIFAR-10/100. 
\begin{figure}[t]
    \centering
    \includegraphics[width=\textwidth]{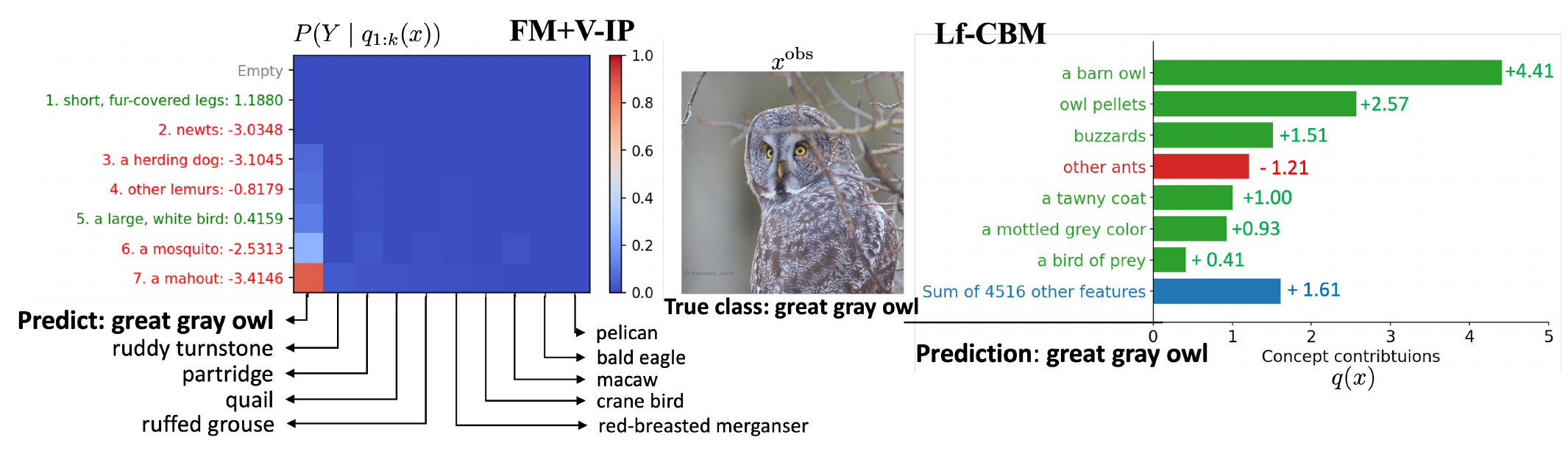}
    \caption{An example of one test sample evaluated using (left) FM+V-IP and (right) Lf-CBM.}
    \label{fig:vip_vs_cbm}
\end{figure}

{
\myparagraph{Comparison between CBMs and V-IP} Fundamentally, CBMs and V-IP are two different explanation methods (Figure~\ref{fig:vip_vs_cbm}). V-IP decomposes a model's prediction by sequentially selecting simpler semantic queries and updating the posterior given the query-answers obtained so far, thereby elucidating the model's reasoning process. In sharp contrast, CBMs find a simple linear mapping which draws correlations between concepts and task variables (classes) while balancing the trade-off between sparsity and accuracy. Ultimately, CBMs and V-IP both offer valid explanations, and the user may interpret predictions based on their preference of frameworks. Further method details can be found in Appendix~\ref{app:method_details}.}

\subsection{Designing the Query Set}\label{sec:design_Q}
We describe the method of generating a query set $Q$ by prompting GPT-3. Then, we compare with three concept filtering methods from Lf-CBM, and argue that queries filtered by these methods are never selected when using V-IP, hence removing the need to do concept filtering for the purpose of V-IP.   

\myparagraph{Query Generation from LLMs} Given that the query set is user-defined, ideally the user is a domain-knowledge expert who is knowledgeable about the discriminative information for a given task. Nonetheless, both manually creating the query set and labeling each sample is a cumbersome and time-consuming task that is infeasbile for large-scale datasets. In this work, we formally extend the V-IP framework by leveraging LLMs to generate interpretable query sets. 

We first assume that for popular image classification tasks in the computer vision community such as ImageNet and fine-grained image classification tasks such as flower classification with Flower-102~\citep{nilsback2008automated}, LLMs contain expert knowledge about both common and fine-grained objects types/classes. Consequently, LLMs would ideally have learned sufficient information about our image classification task to generate a task-dependent, informative and interpretable query set. Moreover, LLMs often come in the format of user-friendly APIs that take in the form of natural human language for both inputs and outputs. The familiarity of objects-in-the-wild, colloquiality and ease-of-access in LLMs make them an attractive choice to extend our query set designs beyond human experts.

To generate interpretable concepts, we prompt GPT-3 to describe class of objects in adjectives, which then can be further processed into phrases or words as concepts. 
This poses the question of how to create an \textit{efficient} prompt that can provide us with a set of interpretable and task-relevant queries. 
In this work, we choose a single all-purpose prompt (with appropriate parameters) as our input to GPT-3:
\begin{quote}
    List the useful visual attributes (and their values) of the \{\texttt{object}\} image category `\{\texttt{class\_name}\}'.
\end{quote}
Here the \{\texttt{object}\} parameter is an argument to the prompt depending on the type of object we are classifying in the dataset. For instance, we replacing \{\texttt{object}\} with ``bird'' for bird classification, with ``scene'' for scene classification, and leave it blank for common object classification such as ImageNet and CIFAR-10/100. Then, for a given dataset, we iterate each class by replace the `{\texttt{class\_name}}' parameter with the name of each class. Take CIFAR-10 as an example, we separately input 10 prompts with `\{\texttt{object}\}' as blank and `\{\texttt{class\_name}\}' as 
``cars'', ``birds'', ``cats'', ``deer'', ``dogs'', ``frogs'', ``horses'', ``ships'', ``trucks''. An example of one run by feeding this prompt to GPT-3 can be found in the Appendix.

\begin{figure}[t]
    \centering
    \includegraphics[width=\textwidth]{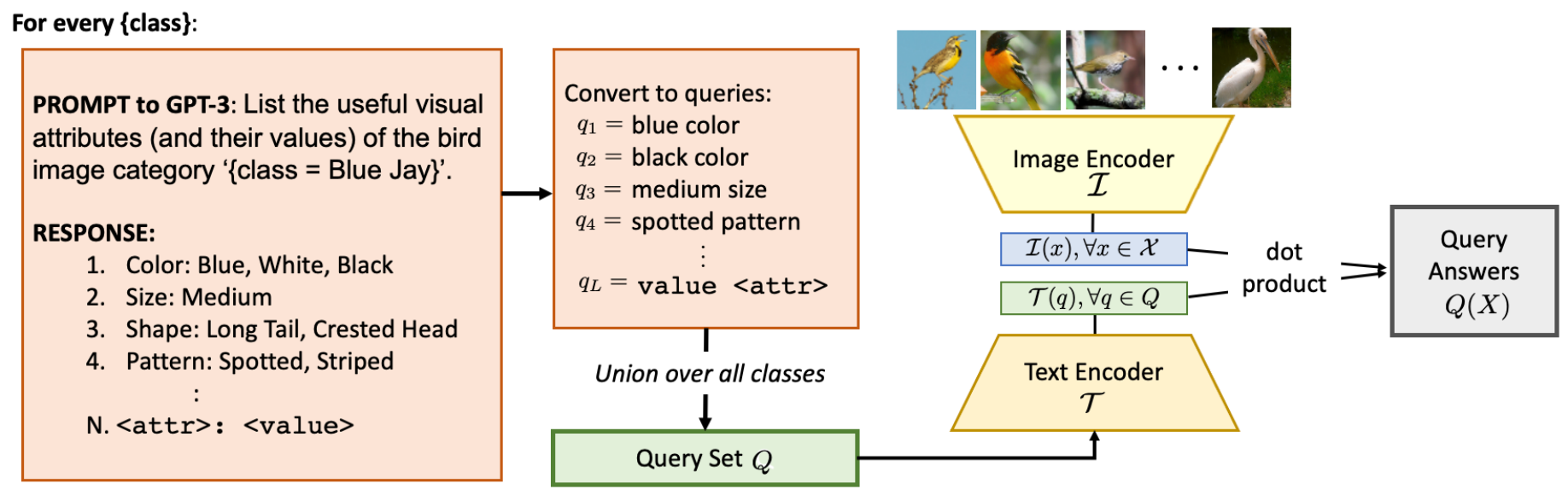}
    \caption{An overview of the method of generating query sets $Q$ and query answers $Q(X)$.}
    \label{fig:overview}
\end{figure}

The output of this prompt is a list, where each item is in the format of \texttt{<attribute>:<value>} (with the possibility of having multiple values). From here, one attribute-value pair can then be transformed as a single query or concept in the format of \texttt{<value> <attribute>}. An example of the entire process is shown in Figure~\ref{fig:overview}, where we prompt for visual attributes and values of a ``Blue Jay''. Specifically, the attribute-value pair ``Size: Medium'' is converted into a query as ``medium size''. Therefore, for a $K$-class classification problem, a GPT-3 generated query set consist of a union of queries transformed from attribute-value pairs output by different $K$ class dependent prompts. In Section~\ref{exp:suff}, we empirically verify that GPT-3 is capable of generating an interpretable and task-dependent query set that is suitable for our tasks. 


\myparagraph{Concept Filtering}
Recent methods such as Lf-CBM\citep{oikarinen2023label} show that concept filtering, removing certain concepts from the initial generated set from GPT-3, is necessary to achieve good test performance and interpretability. For comparison, we look at some steps of concept filtering in Lf-CBM. We argue that, in the context of the IP algorithm in ~\ref{alg:IP}, filtered queries/concepts are uninformative, hence will naturally not be selected by our formulation. For context, we consider $N$-class bird classification as our task and context, and again denote $x$ and $y$ as the realizations of the data and label random variable $X$ and $Y$, and $S$ as history.
\begin{enumerate}
    \item \textit{Concept Filter 1: Delete concepts too similar to task's class names.} Consider the binary query: $q = $ ``Is the bird in the image a Blue Jay?''. From the definition of mutual information between $q(X)$ and $Y$, $I(q(X), Y) = H(Y) - H(Y \mid q(X)) = H(Y \mid q(X) = 1)P(q(X) = 1) + H(Y \mid q(X) = 0) P(q(X) = 0).$ Since the answer to the query is Yes, i.e. $q(X) = 1$, there is no ambiguity left in $Y$ and $H(Y \mid q(X) = 1)P(q(X) = 1) = 0$. Hence, $I(q(X), Y) = H(Y \mid q(X) = 0) P(q(X) = 0).$ Now, if $Y$ follows a uniform distribution, then $H(Y) = \log N$ and $H(Y \mid q(x) = 0) P(q(X) =0 ) = (1 - 1 / N)\log(N - 1)$. For a large $N$, $H(Y) \approx H(Y \mid q(X) = 0)P(q(X) = 0)$, implying $I(q(X), Y) = H(Y) - H(Y \mid q) \approx 0$. Therefore, $q$ is an uninformative query and will not be selected.
    
    \item \textit{Concept Filter 2: Delete concepts too similar to other concepts.} Consider $q_1 = $ ``Is this bird tiny?'' and $q_2 = $ ``Is this bird small?''. Reasonably, $q_1$ and $q_2$ provide similar information. Since the IP algorithm selects queries in a sequential manner, once $q_1$ has been chosen, $q_2$ is unlikely to be the next most informative query. 
    Take the extreme case where $q_1(x) = q_2(x)$ for all $x \in \mathcal{X}$, the mutual information between labels $Y$ and $q_2(X)$ is exactly 0 when $\{q_1, q_1(X)\} \in 
    S$, i.e. $I(Y ; q_2(X) \mid S,  q_1(X)) = 0$. 
    \item\textit{Concept Filter 3: Delete concepts not present in training data.} Consider a query $q = $ ``Is the car red?''. In a bird classification task, a car is irrelevant for the task, hence $q(x) = 0$ is constant for all $x$. This also implies $I(Y ; q(X) \mid S) = 0.$ 
\end{enumerate}

While concept filtering is not needed for V-IP to select the most informative query, since V-IP requires differentiation through a high-dimensional vector of query logits, filtering can reduce the size of $Q$, making V-IP more computational efficient. Although in our experiments we see little-to-no difference in time to train, finding ways to reduce the size of $Q$ can be useful for applications such as extreme classification problems~\citep{daghaghi2021tale}.

\subsection{Answering Queries}\label{sec:qry_ans}
Once a query set $Q$ is defined, we require a method of computing the query answers $q(x)$ for every $x \in \mathcal{X}$. Previously, query answers are either annotations of the dataset or feature values such as pixel values in a given image. In this work, our generated query sets are concepts expressed in natural human language and we will leverage CLIP to generate query answers $q(x)$.

CLIP is a vision-language model trained on image-texts pairs in a contrastive manner, in which the inner product (with range $[0, 1]$) of the output image and text embedding represents the similarity of a given input image and text. CLIP consists of two encoders: an image encoder $\mathcal{I}(\cdot)$ and a text encoder $\mathcal{T}(\cdot)$, where $\mathcal{I}$ takes images $x$ as input and outputs image vector embeddings $\mathcal{I}(x)$, whereas $\mathcal{T}$ takes text $q$ (variable size sentences) as input and output a text embedding $\mathcal{T}(q)$ as output. Note that the dimensions of both $\mathcal{T}(x)$ and $\mathcal{I}(q)$ are the same, and we assume all embeddings are $\ell_2$-normalized.

Since the dot-product $\mathcal{I}(x) \cdot \mathcal{T}(q)$ encodes the similarity between the given image $x$ and query $q$, we compute the query answer $q(x)$ for all $q \in Q$ and $x \in \mathcal{X}$ as the raw dot-product (score) between the image embedding $\mathcal{T}(x)$ and text embedding $\mathcal{I}(q)$, i.e. $q(x) = \mathcal{I}(x) \cdot \mathcal{T}(q).$
When $q(x)$ is large (close to 1), the query $q$ and the image $x$ are similar, whereas when $q(x)$ is small (close to 0) they are dissimilar. Following Lf-CBM, we also Z-score standardize each query answer by subtracting the mean and dividing the standard deviation of all query answers from the training set.

In \citet{chattopadhyay2022interpretable, chattopadhyay2023variational}, queries are binary questions about the data such as "Are there square windows in the building?". In contrast, in our framework query answers measure the similarity between the concept and the data, e.g., the similarity between the concept "Square windows" and an image of a building. Answering queries in the form of questions is related to the field of Visual Question Answering (VQA). 
We reserve the research of more sophisticated and fine-grained queries for future work.

\section{Experiments}\label{section: Experiments}
In this section, we empirically demonstrate that by leveraging LLMs and VLPs, FM+V-IP addresses the key challenges for doing interpretable-by-design predictions in three parts: First, we show that LLMs are capable of generating interpretable and informative query sets (Section~\ref{exp:suff}). Then, we show that V-IP can achieve competitive test performance with both filtered and unfiltered query sets when compared to Lf-CBM (Section~\ref{sec:compare-with-lfcbm}). Finally, we show that FM+V-IP with large query sets can achieve comparable test performance with much fewer number of queries when compared to LaBo (Section~\ref{sec:compare-with-labo}). In the Appendix, we also showcase an example where a sample is originally misclassified, then later intervening and corrected by modifying an erroneous query answer. Last but not least, we document the details of our experiments, such as training procedure, architecture design, hyperparameters for FM+V-IP, Lf-CBM and LaBo, as well as examples of queries from each query set and test samples in the Appendix. Code is available at \url{https://github.com/ryanchankh/FM-V-IP/}.

\begin{figure}[t]
  \centering
  \includegraphics[width=\textwidth]{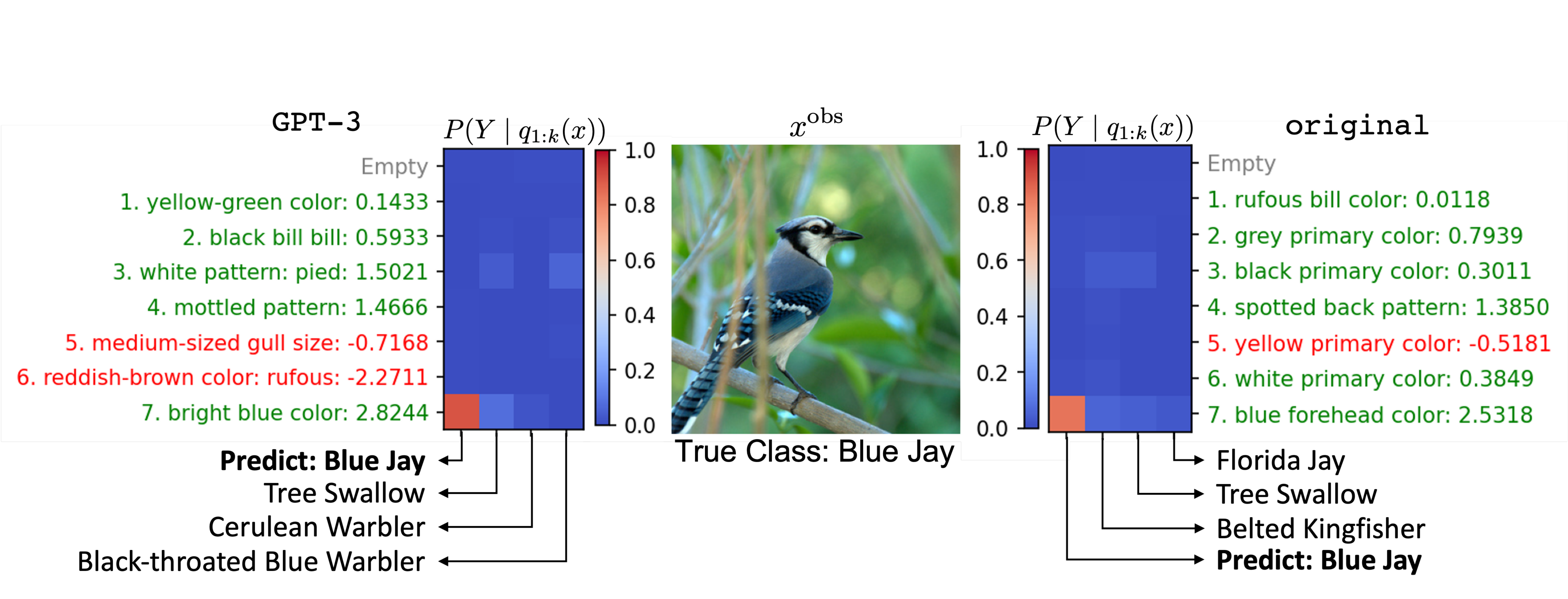}
  \caption{Comparisons of query-answer chains using V-IP models trained with  (\textbf{left}) \texttt{GPT-3} query set and (\textbf{right}) \texttt{original} query set. Each $k$-th row corresponds to the posterior after selecting $k$ queries. In the $x$-axis, only the four classes with the highest posterior probabilities at the stopping criteria are shown, with the predicted class in bold. Green and red $y$ axis labels corresponds to whether $q_k(x)$ is above or below 0, with their query answer $q_k(x)$ shown next to $q_k$. }
  \label{fig:traj_suff_compare1}
\end{figure}

\subsection{Evaluating Query Sets from LLMs with Annotated Data}\label{exp:suff}

We first demonstrate that GPT-3 is capable of generating an interpretable and informative query set. We construct two different query sets for bird classification task using CUB-200~\citep{wah2011caltech}. The first query set, named \texttt{original}, has a size of 312 and is constructed from the original 312 attributes of the CUB-200 dataset, equivalent to the query set used for V-IP in previous works~\citep{chattopadhyay2022interpretable}. The second query set, named \texttt{GPT-3}, has a size of 518 and is constructed from concepts obtained using the single-design prompt described in Section~\ref{sec:design_Q}.
The query answers $q(x)$ are generated using CLIP dot-product as mentioned in Section~\ref{sec:qry_ans}.We train one V-IP model for each of the query set and compare their test performance. 

\begin{wrapfigure}[14]{r}{0.35\textwidth}
  \centering   
  \vspace{-2mm}
  \includegraphics[width=\textwidth]{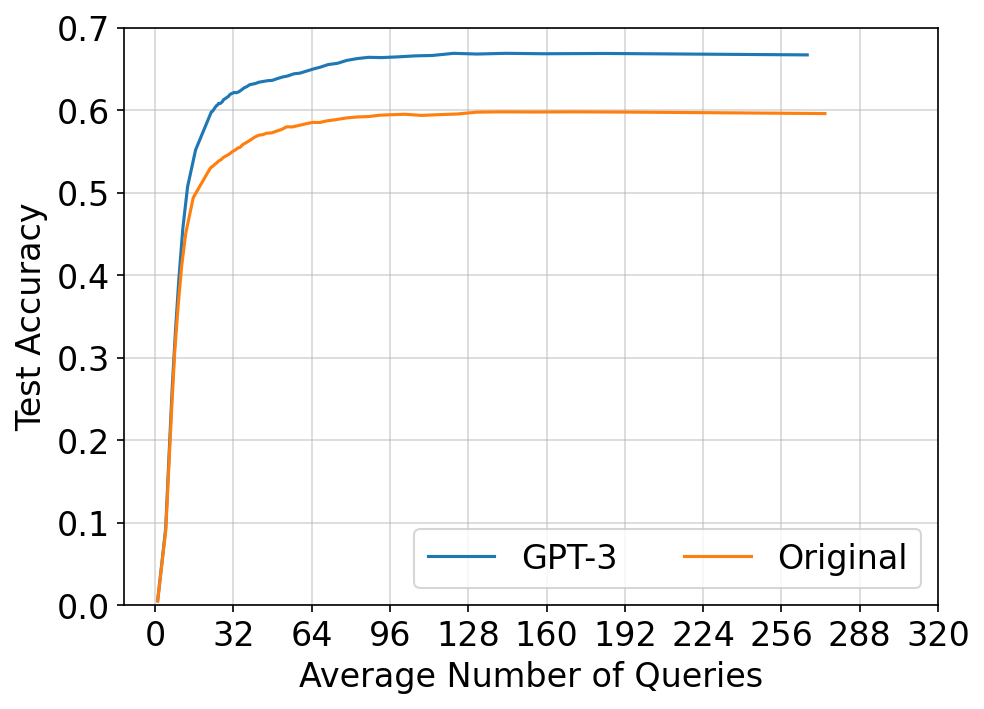}
      \caption{\small{Test performance of CUB-200 using the original query set \texttt{original} versus GPT-3 generated query set \texttt{GPT-3}. }}\label{fig:suff}
\end{wrapfigure}

By varying $\epsilon$ and evaluating at different stopping criteria $P(Y \mid q_{1:k}(X)) > 1 - \epsilon$, we compute the average number of queries needed for each test sample, and compute the trade-off between test accuracy and average number of queries used in Figure~\ref{fig:suff}. We observe that the test accuracy gained by increasing the number of queries is similar before the 16$^\text{th}$ query, but as the number of queries increases, V-IP performs approximately 7\% better with the \texttt{GPT-3} query set than with the \texttt{original} query set after using 96 queries. This implies that the \texttt{GPT-3} query set contains additional queries that are more informative for predicting the bird class than the \texttt{original} queries. We demonstrate this in a direct comparison of query-answer chains shown in Figure~\ref{fig:traj_suff_compare1}: Note that queries related to colors are mainly selected from the \texttt{original} query set, whereas queries related to different bill shapes and patterns such as ``white pattern: pied'' and ``mottled pattern'' are selected from the \texttt{GPT-3} query set. These queries help further distinguish birds with similar visual features. We show more examples of query-answer chains using the two query sets in the Appendix.

\subsection{Evaluating Filtered and Unfiltered Query Sets with Lf-CBM}\label{sec:compare-with-lfcbm}

We use Lf-CBM as our baseline and evaluate the test performance of V-IP using the filtered and unfiltered query/concept set as Lf-CBM on five datasets: CIFAR-10 and CIFAR-100~\citep{krizhevsky2009learning}, ImageNet~\citep{deng2009imagenet}, CUB-200~\citep{wah2011caltech}, and Places365~\citep{zhou2017places}. Label-free CBM (Lf-CBM)~\citep{oikarinen2023label} is an extension of CBM, where concepts/queries $Q$ are first generated by prompting GPT-3, then refined by passing through a series of concept filters, such as those mentioned in (Section~\ref{sec:design_Q}). Given the query answers generated from image embeddings of pre-trained neural networks, a linear predictor is trained using an elastic-net objective with sparsity parameter $\lambda$ to map from concept scores to class logits. Note Lf-CBM differs from V-IP in our experiments in two ways:
1) The predictor for Lf-CBM is linear, whereas the predictor for V-IP is a nonlinear neural network; and 2) The number of concepts used from Lf-CBM is decided by the sparsity parameter $\lambda$, whereas queries used per sample in V-IP is variable-length and driven by mutual information. An example of the two methods is shown in Figure~\ref{fig:vip_vs_cbm}. We give the full explanation and details of the Lf-CBM in the Appendix. For comparison, we denote our method as FM+V-IP. 

\begin{figure}[t]
  \includegraphics[width=0.85\textwidth]{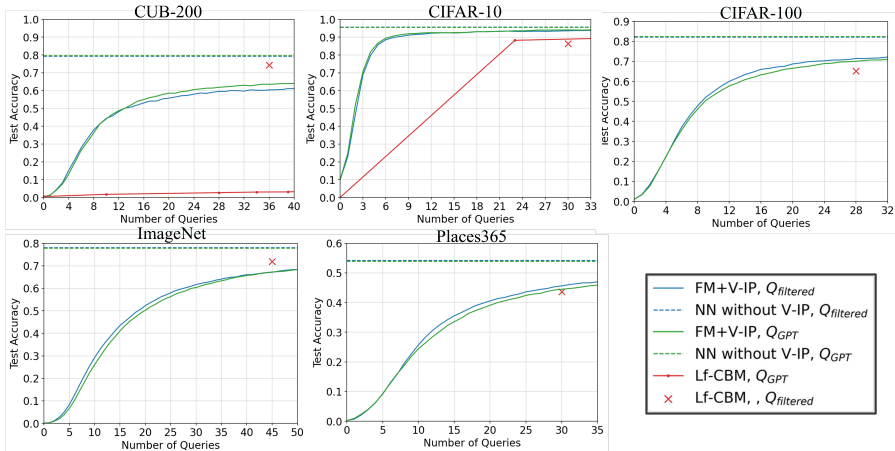}
  \caption{Test performance of FM+V-IP versus Lf-CBM on CIFAR-10, CIFAR-100, CUB-200, ImageNet, and Places365. Using a \textit{filtered} query set $Q_{\text{filtered}}$, blue solid lines are results from FM+V-IP and blue dashed lines are baseline results from a neural network classifier without V-IP.  Using an \textit{unfiltered} query set $Q_{\text{GPT}}$, green solid lines are results from FM+V-IP and green dashed lines are baseline results from a neural network classifier without V-IP. Red solid lines are results by training multiple Lf-CBMs with $Q_{\text{GPT}}$ and varying the sparsity parameter, while red crosses are results from Lf-CBMs with $Q_{\text{filtered}}$, directly referenced from \citet{oikarinen2023label}.}\label{fig:vip_vs_lfcbm}
\end{figure}
\begin{wrapfigure}[32]{r}{0.3\textwidth}
  \centering
  \vspace{-5mm}
  \includegraphics[width=\textwidth]{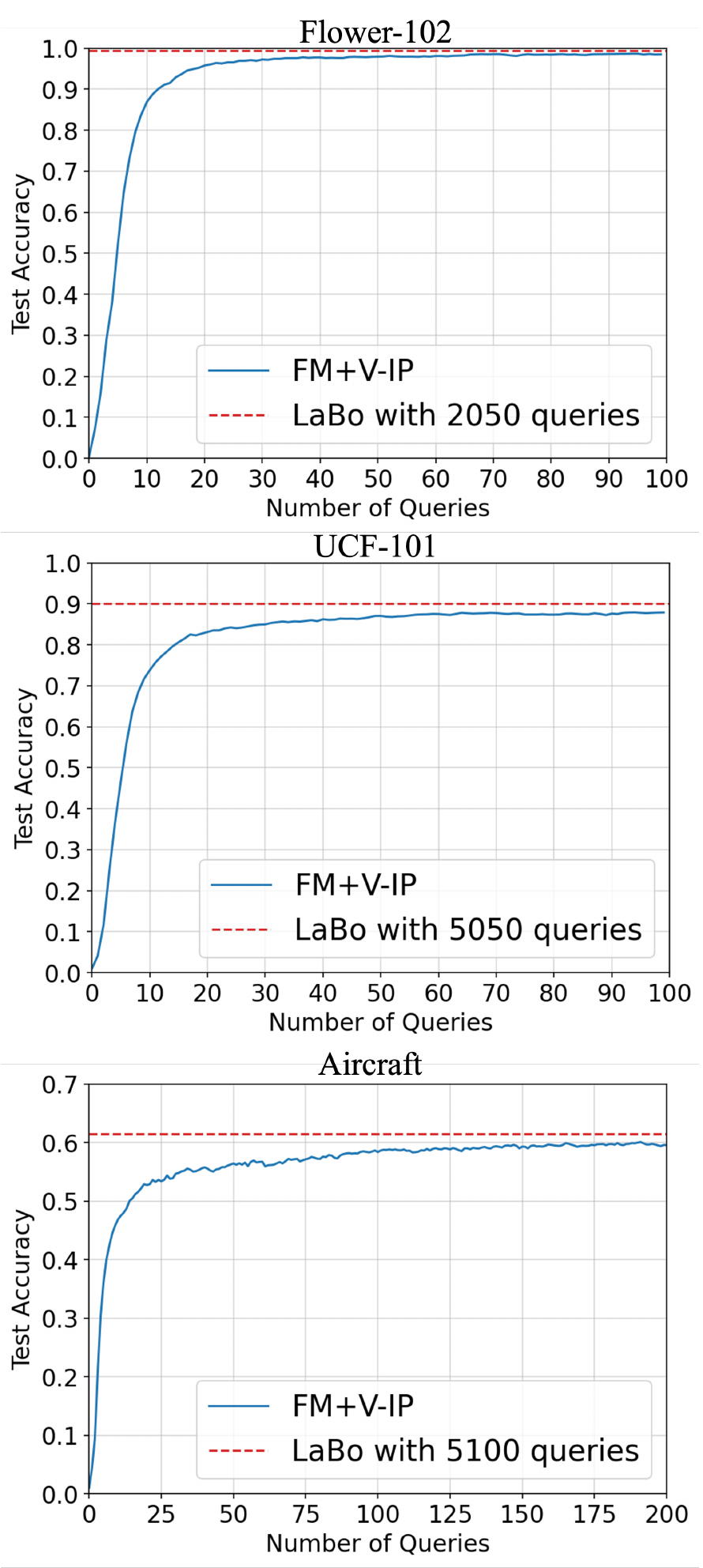}
  \caption{\small{Comparison of test performance between FM+V-IP and LaBo. Red curve represents the test accuracy of LaBo after a certain number of queries. The blue curve represents the test accuracy of FM+V-IP at varying number of queries.}}
  \label{fig:labo-compare}
\end{wrapfigure}

We compare the test performance of FM+V-IP using unfiltered query set $Q_{\text{GPT}}$ from our single all-purpose prompt to GPT-3 versus Lf-CBM using a filtered query set $Q_{\text{filtered}}$ from Lf-CBM. In Figure~\ref{fig:vip_vs_lfcbm}, we demonstrate that FM+V-IP does not require any concept filtering to achieve comparable test performance. Take CIFAR-10 as an example: using with 30 queries, V-IP achieves 94.11\% test accuracy, whereas Lf-CBM achieves 86.40\% test accuracy. From an alternative viewpoint, FM+V-IP achieves the same test accuracy as Lf-CBM using fewer number of queries. For instance, FM+V-IP only requires 5 queries to achieve the 86.84\% test accuracy whereas Lf-CBM requires 30 concepts to achieve similar test accuracy. On the other hand, for ImageNet, FM+V-IP obtains a lower test accuracy than Lf-CBM using 45 queries, where the former achieves a test accuracy at 67.11\% and latter at 71.95\%. This can be attributed to the fact that query answers generated by CLIP are too noisy for the given query sets. Similarly for CUB-200, FM+V-IP achieves a 63.55\% test accuracy whereas Lf-CBM achieves a test accuracy of 74.31\%. The drop in performance is due to the fact that CLIP dot-products are not ideal for representing fine-grained details of the image (as mentioned in Section~\ref{sec:qry_ans}).

Moreover, in Figure~\ref{fig:vip_vs_lfcbm}, we compare the performance of FM+V-IP and Lf-CBM on the same query set $Q_{\text{GPT}}$ without any filtering. Similar to findings in Lf-CBM, we run into memory issues when trying to experiment with larger datasets. Therefore, we only report findings on datasets where the size of the unfiltered query set is feasible for performing this experiment. For CIFAR-10 and CUB-200, we train multiple-Lf-CBMs by varying the sparsity parameter $\lambda$ and compute the test accuracy for different number of queries. On the other hand, we train a single V-IP model for each dataset. For CIFAR-10, test accuracy for FM+V-IP saturates around 95\% using 25 queries, whereas test accuracy for Lf-CBM never achieves 95\%, and requires 80 queries to achieve 92\% test accuracy. For CUB-200, FM+V-IP achieves 65\% with 50 queries, whereas Lf-CBM fails to generalize well. This shows that FM+V-IP is feasible to train and can perform well with any query set. On the other hand, Lf-CBM requires concept filtering to become feasible or perform well. 

Last but not least, while $Q_{\text{GPT}}$ contains more uninformative queries than $Q_{\text{filtered}}$, Figure~\ref{fig:vip_vs_lfcbm} also shows that FM+V-IP still selects informative queries and attains similar test performance at different number of queries. 

\subsection{Comparing V-IP and LaBo with the same query set}\label{sec:compare-with-labo}
While V-IP optimizes to find short variable-length query-chains, LaBo first optimizes a submodular function to select a pre-determined number of suitable concepts for each class, then predicts using a linear classifier. In this section, we compare the test performance of V-IP versus LaBo using the same query set on three datasets: Flower-102~\citep{nilsback2008automated}, UCF-101~\citep{soomro2012ucf101} and FGVC-Aircraft (Aircraft)~\citep{maji13fine-grained}. For a given dataset, the query set is generated by feeding multiple prompts to GPT-3 to obtain sentence descriptions for each class, then processed into individual concepts using the LLM T-5~\citep{raffel2020exploring}. In \citet{yang2022language} and in this experiment, the query answers are dot-products of CLIP embeddings between concepts as text and images. We train one V-IP model for each dataset with their respective query sets generated from LaBo. 

In Figure~\ref{fig:labo-compare}, we show that, while both methods do not require any concept filtering, V-IP requires 0.02-0.05 times the number of queries of LaBo to achieve comparable performance. The size of the query set $Q$ are orders of magnitude larger than filtered concept sets in Lf-CBM: 24960 for Flower-102, 46845 for UCF-101 and 36324 for Aircraft. This further argues that V-IP can perform \textit{and scale} competitively with large query sets without requiring downsizing of the query set.

\section{Conclusion}\label{section: Conclusion}


In this work, we address the need for data with concept annotations by proposing a method of leveraging LLMs and VLPs to create query/concept sets and annotations. We mathematically argue that the sequential selection process of V-IP avoids the need for any concept filtering, and empirically show that V-IP is capable of achieving competitive test accuracy on different scales of datasets when compared with other methods such as Lf-CBM and LaBo. Nonetheless, we also observe that using CLIP dot-products as query answers can be noisy for fine-grain concepts, negatively impacting test performance. As future work we seek to extend the framework for more sophisticated queries and answering them with more scalable and precise VQA systems, so that task predictions can obtain better accuracy and be more interpretable. 

\bibliographystyle{plainnat}
\bibliography{references.bib}

\newpage
\appendix
\section*{Appendix}\label{sectopm: Appendix}
\section{Extended Results}
\subsection{Intervention on Misclassified Examples}

One key contribution from CBMs from \citet{koh2020concept} is the ability to intervene a model's decisions. This ability is especially useful in the case where the classifier make a erroneous prediction due to a mis-labled concept. In this section, we demonstrate one example of intervention on models trained with FM+V-IP.

While in this work we mainly focus on using dot-products of CLIP image and text embeddings as query answers, since CLIP is not trained for the purpose of VQA, query-answers can be noisy and incorrect. Hence, the ability to interpret and intervene a model's decisions has great significance, especially erroneous prediction due to an incorrectly annotated concept/query-answer.

\begin{figure}[h]
  \begin{center}
    \includegraphics[width=0.75\textwidth]{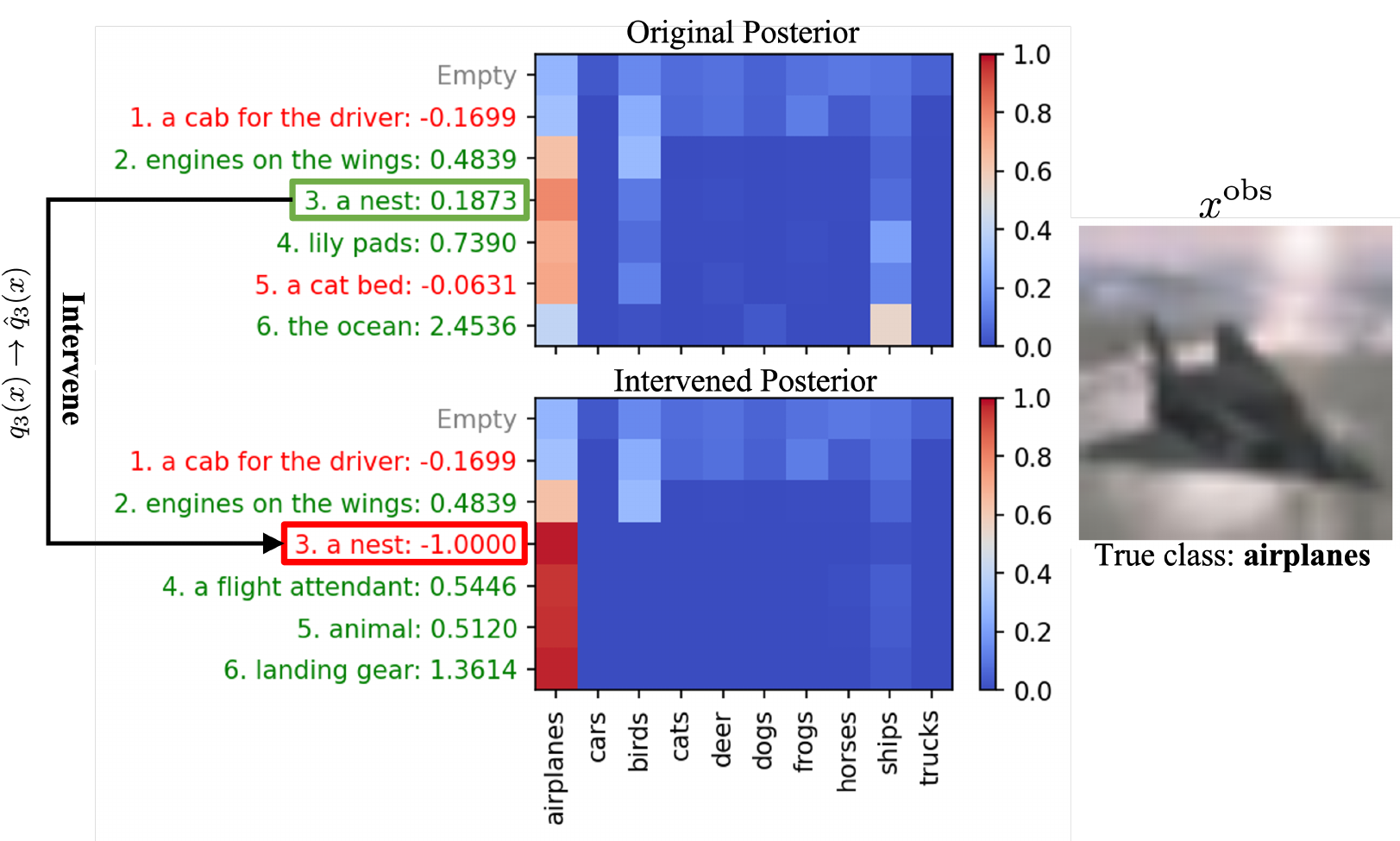}
    \caption{Intervention on an misclassified sample in CIFAR10.}\label{fig:intervene}
  \end{center}
\end{figure}
Here we illustrate one example that FM+V-IP has the ability to intervene the sequential selection process of queries during inference and correct a misclassification result. A misclassified image of an airplane from CIFAR-10 is shown in Figure~\ref{fig:intervene}. The top figure shows the posterior using original query-answer chain (history) $P(Y \mid S = q_{1:6}(x))$, making an incorrect prediction of the class ``ship''. Note that since the object in the image does not resemble a (bird) nest, they should be dissimilar and have a negative score. Hence, following the sequence of queries selected by the trained model, we edit the query answer $q_3(x) \rightarrow \hat{q}_3(x)$ for ``a nest'' from a positive score 0.1873 to a negative score $-1$,  With the updated query answer $\hat{q}_3(x)$, the model now outputs the correct posterior $P(Y \mid S \setminus q_3(x), \hat{q}_3(x))$ and predicts the correct class label ``airplanes'' with high confidence.

\subsection{Test Performance of FM+V-IP with More Queries}
\begin{figure}[h]
    \centering
    \includegraphics[width=\textwidth]{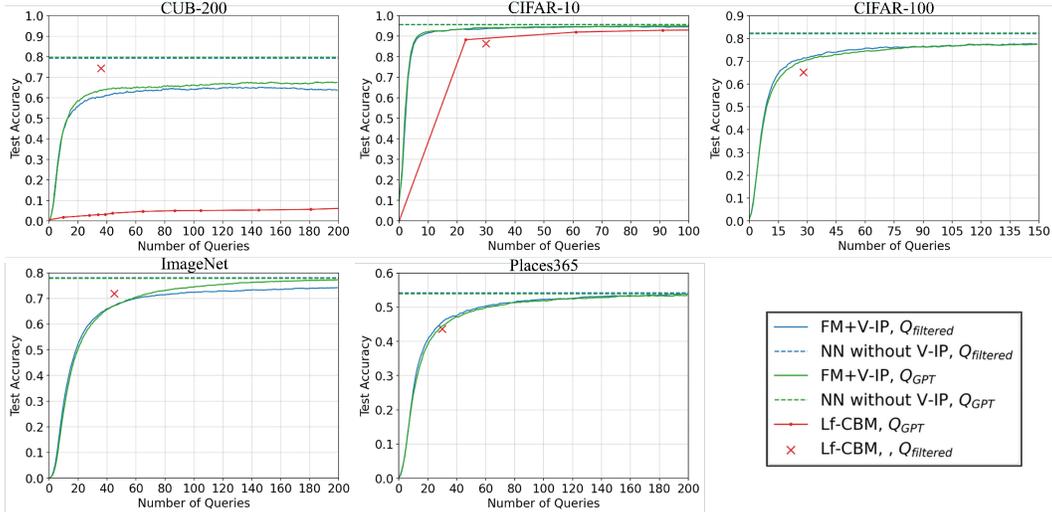}
    \caption{Test performance of FM+V-IP and Lf-CBM. This figure is the same figure as Figure~\ref{fig:vip_vs_lfcbm}, except each model is evaluated with longer number of queries. }
    \label{fig:vip_vs_lfcbm_full}
\end{figure}

We show extended results of test performance of FM+V-IP versus Lf-CBM on different datasets and query sets in Figure~\ref{fig:vip_vs_lfcbm_full}. For any dataset, the baseline result is computed by training only the classifier of the FM+V-IP model on all query answers; This is equivalent to training a vanilla fully-connected neural network with all query answers as input features. Ideally, the test performance of the baseline classifier should approximately upper bound the test performance of models trained with FM+V-IP with any number of queries. 

From Figure~\ref{fig:vip_vs_lfcbm_full}, we see that as the number of queries increase, the test performance of FM+V-IP approaches the baseline test performance. Specifically, we see that CIFAR-10, ImageNet and Places365 can achieve the same test performance as baseline at 100, 200, and 200 queries respectively. Nonetheless, we also observe that CUB-200 and CIFAR-100 never arrives at the desired test performance, even with extra number of queries. This could be attributed to the expressivity of the architecture. In FM+V-IP, the choice of the predictor $f_\theta$ is arbitrary and flexible. Hence, the gap in performance could potentially be addressed via better neural network architecture search.

{
\subsection{Accuracy of Attributes Evaluated on Annotated Dataset}
Here we evaluate the accuracy of CLIP-generated answers (i.e. $q(x) = \mathcal{I}(x) \cdot \mathcal{T}(q) \in [0, 1]$). Specifically, using the CUB-200 dataset, we compare the answers obtained from CLIP and the expert-labeled annotated attributes from CUB-200. Since given attributes are binarized, we use the validation set to compute the best threshold for binarization, where $q(x)$ rounds up to 1 if it is above the threshold, and rounds down 0 if $q(x)$ is below the threshold. For each train/test/validation split, we first compute the accuracy for each attribute over all samples, then compute the average attribute accuracy by taking the average of accuracies over all attributes. For the training set, we obtain an average accuracy of 88.3\%. For the validation and test set, both obtain an average accuracy of 88.2\%. We highlight that although the query answers are not fully correct, noisy query answers do not have a large impact on the interpretability of FM+V-IP. 
}

{ 
\subsection{Explanations from FM+V-IP}
For each input, our method FM+V-IP generates a query-answer chain as a means of explaining the model's decision making, breaking it down into simpler questions and provide a sequential reasoning to the prediction problem. We explicitly showcase this reasoning in words in Figure~\ref{fig:app-explain}. This coarse-to-fine reasoning is vastly different than explanations that CBMs offer, which is to find a sparse set of concepts that most correlate to the input, for prediction. We hope to highlight the benefits of the use of foundational models, including concept filtering via mutual information, with V-IP through this work.

\begin{figure}[h]
    \centering
    \includegraphics[width=0.75\textwidth]{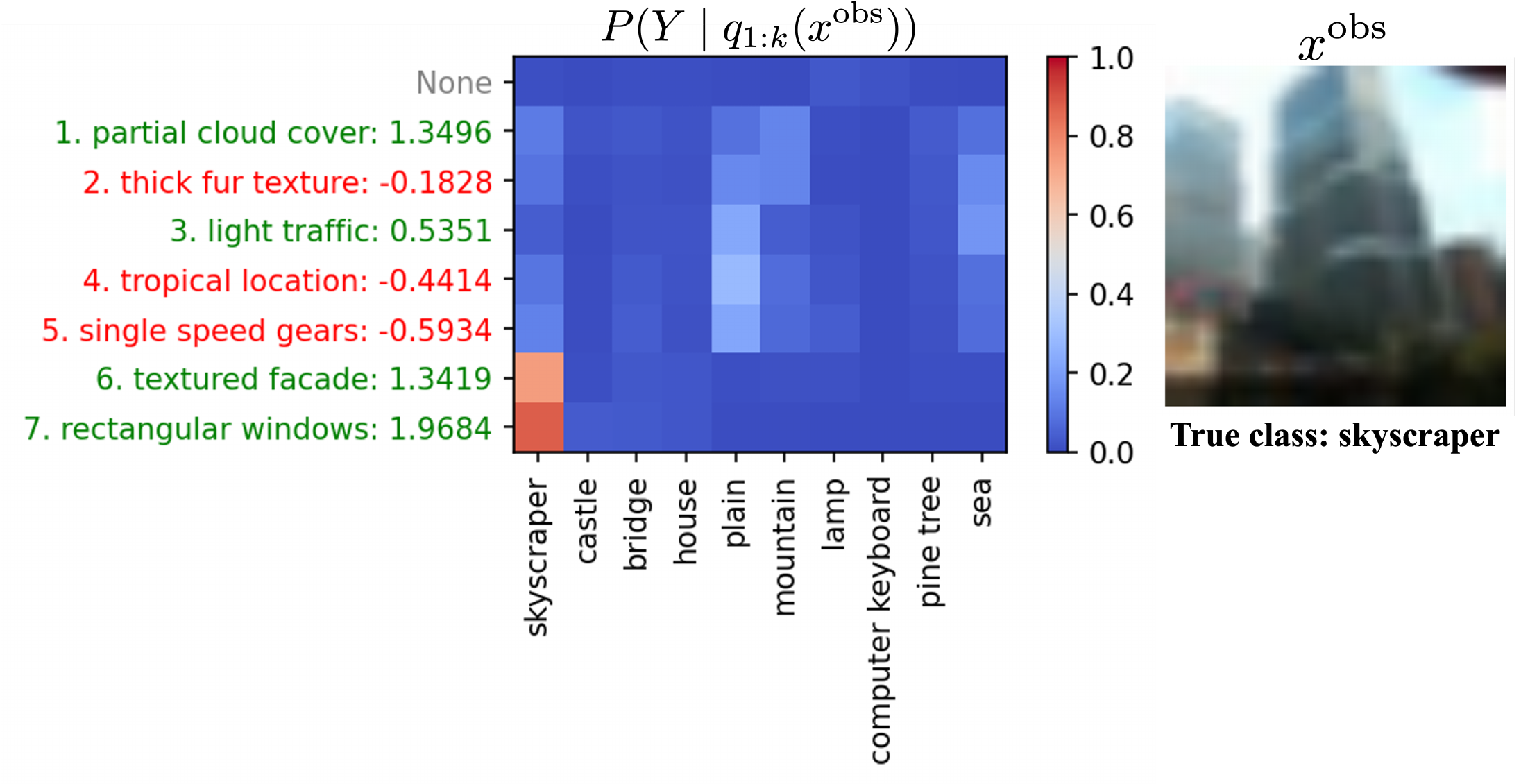}
    \caption{An example of query-answer chain from CIFAR-100. FM+V-IP first selects queries such as “partial cloud cover” and “thick fur texture” to differentiate between the “animals” super-classes and “buildings” super-classes. It further ask queries such as “light traffic” to differentiate between indoor and outdoor buildings. Finally, it ask queries such as “rectangular windows” to differentiate the finer classes, namely as skyscraper, castle and bridges. }
    \label{fig:app-explain}
\end{figure}
}

\section{Limitations}
In this work, we only focus on the task of interpretable image classification using FM+V-IP with GPT-3 for generating queries and CLIP dot-products for answering. One limitation of FM+V-IP is that CLIP dot-products can be noisy and inaccurate, which can severely impact predictive accuracy. Reasonably, CLIP is pretrained on large number of text and image pairs, which the text may or may not describe the image in a fine-grain enough detail to accurate represent any data. A future work may involve using Visual Question Answering (VQA) systems to answer queries, which allows the design of queries to be more flexible and the answer of the query to be precise. Ideally, this would not only improve the test performance due to cleaner query answers, but also improve the interpretability of the model.

{ Another limitation is that concepts generated by GPT-3 could potentially be non-visual. While we construct our prompt by explicitly mentioning ``visual attributes," there is no explicit filter that controls the interpretability of the output itself. We reserve incorporating such control in our concept/language generation process~\citep{wu2009perceptual} or using prompt learning for finding more interpretable concepts~\citep{zang2022unified} for future work.}

{In addition, in this work we focus on solving visual tasks, but our framework can be further extended to other modalities including medical data, where generative models such as GPT finetuned on medical data~\citep{haupt2023ai} can be leveraged for generating and answering queries. We also reserve this for further work. }

\section{Broader Impacts}
Intepretable predictions plays a key role in gaining the trust of the user when machine learning models are deployed in real-world uses. In this work, we advocate that interpretability should be a built-in functionality of the models. While this work focuses on common object image classification tasks, our framework can be expanded to other risk-sensitive applications domains such as medical imagining. Meanwhile, explanations has a potential to de-bias algorithms by realizing why the predictor makes a correct or incorrect prediction, and we aim to make machine learning algorithms more accessible and fair for people regardless of their ethnic or social status.

\section{Method Details}\label{app:method_details}
\subsection{Label-free CBM (Lf-CBM)}\label{app:method-lfcbm}
Label-free Concept Bottleneck Model (Lf-CBM) was introduced in \citep{oikarinen2023label} as a method to address the lack of concept-annotated data for making interpretable predictions. For a given image classification task, similar to FM+V-IP, the method first create an initial concept set by prompting GPT-3 to describe different classes from the task. Then, using a series of concept filtering steps, such as removing concepts too similar to class names, or concepts too similar to each other, a concept set is generated. 

The method then learns a weight matrix that projects image embeddings from deep networks into a concept space. Given a pre-trained deep network $f$, such as ResNet-18~\citep{he2016deep}, we first compute an image embedding $f(x)$ for an input image $x$. Then we train a linear network with weight $W_c$ that maps from image embedding to a feature space known as ``concept bottleneck layer'', i.e. $f_c(x) = W_c f(x)$. This layer is trained with a ``cosine cubed'' similarity loss~\citep{oikarinen2023label}, where $f_c(x)$ are encouraged to align with a matrix $P$, constructed from taking the dot-product between image embeddings of all training images and text embeddings of all concepts. Finally, it learns a linear predictor  $g(x) = W_F f_c(x) + b_F$ that predicts the label $y$ of the given image $x$. $g$ is trained with an elastic-net objective:
\begin{equation*}
    \min_{W_F, b_F} \sum^N_{i=1} \mathcal{L}_{\text{CE}} (W_F f_c(x_i) + b_F, y_i) + \lambda R_\alpha (W_F),
\end{equation*}
where $R_\alpha(W_F) = (1 - \alpha) \frac{1}{2} \| W_F \|_F + \alpha  \| W_F \|_{1, 1}$. $N$ is the number of training samples, $\| \cdot \|_F$ here denotes the Frobenius norm and $\| \cdot \|_{1, 1}$ denotes the element-wise 1-norm. $\lambda$ is a parameter to tune for sparsity, the number of concepts used for prediction. In our experiments in Section~\ref{sec:compare-with-lfcbm}, we tune $\lambda$ to achieve different levels of sparsity.

\subsection{Language in a Bottle (LaBo)}
Introduced in \citet{yang2022language}, LaBo also proposes to perform interprertable predictions with CBMs by using LLMs to generate concepts and computing concept scores with dot products of image and text embeddings from CLIP for each image and concept. The method of generating a concept set is as follows: we first prompt GPT-3 to describe attributes of different image classes using various general template. Then, we use a T5 concept extractor~\citep{raffel2020exploring}, another LLM finetuned on extracting key phrases from sentences, to extract short concepts. Finally, concepts with class names are processed, by either replacing the class name with its superclass name or completely removing the concept from the concept set. 

Given $N$ samples of data and labels  $\{x_i, y_i\}_{i=1}^N$ for a $K$-class classification problem, and a candidate set of concepts $S = S_1 \cup \ldots \cup S_K$ obtained from the above method, LaBo optimizes the following composite objective:
\begin{equation*}
    \min_{f, C} \sum^N_{i=1}\mathcal{L}_{\text{CE}}(f(g(\mathcal{I}(x_i), E_C), y_i)) - \mathcal{F}(C, \{\mathcal{I}(x_i), y_i\}_{i=1}^N).
\end{equation*}
The above adjective is solved sequentially: first we solve for the maximization problem of seeking the optimal set of concepts $C$ from a large candidate set $S$, then solve the minimization of cross entropy loss for the linear predictor $f$. Again, $\mathcal{I}$ denotes the CLIP image encoder and $\mathcal{T}$ denotes the CLIP text encoder.

Noteably, $\mathcal{F}$ is a submodular function that defines a discriminability score and a coverage score for any subset $C$. Maximizing $\mathcal{F}$ leads to an optimal subset of concepts $C_y$ from a large candidate set $S_k$ for each class that optimizes for finding discriminative concepts (similar to not having two concepts that similar to each other) and concepts that emcompass most of the semantics of the overall candidate concept set.

Having solved the maximization problem, $C$ determines the choice of $E_C$, which is a matrix whose row vectors are text embeddings of all concepts $c \in C$, i.e. $E_C = [\mathcal{T}(c_1) | \ldots | \mathcal{T}(c_{|C|})]^\top$. And $g(x, E_C) = E_C x$ is a matrix multiplication of the image embedding $x$ with text embeddings of selected concepts, hence the output of $g$ is a vector of concept scores for $x$. Finally, similar to CBMs, $f$ consisnt of a weight matrix $W$ that maps concept scores to label logits. 


\subsection{Difference between Lf-CBM, LaBo and FM+V-IP}
We emphasize that all three methods, Lf-CBM, LaBo, and FM+V-IP, use GPT-3 as a starter to generate concepts. Nonetheless, our work advocates that FM+V-IP do not require any type of concept filtering, whereas Lf-CBM requires multiple step of ad-hoc concept filtering and LaBo requires first finetuning a T5 LLM and similar concept filtering steps. Not having the need for concept filtering, either for the sake of improving test accuracy or interpretability, is the main advantage of FM+V-IP, as removing concepts involves the risk of discarding potentially meaningful concepts. 

In addition, both LaBo and FM+V-IP use dot-products of image and text embeddings from CLIP directly as concept scores/query answers, whereas Lf-CBM trains a concept bottleneck layer that maximizes alignment between transformed image embedding ($f_c(x)$ from Appendix Section~\ref{app:method-lfcbm}) and CLIP dot products. From our experiments, we see that CLIP dot-products are generally suitable for our image classification tasks, but image embeddings may be fetter fine-grain classification, since pre-trained models are finetuned on images from the given dataset of the task for extracting fine-grain features. 

Finally, the number of queries/concepts for each prediction is chosen differently: In Lf-CBM, the sparsity parameter $\lambda$ decides the number of concepts used, where a larger $\lambda$ leads to smaller number of concepts. In \citet{oikarinen2023label}, $\lambda$ is chosen such that the model only uses 25-35 concepts. In LaBo, each classes selects a pre-defined number of concepts. The user has control over this parameter and can chooses exactly however many concepts LaBo uses for prediction. In our work FM+V-IP, the number of queries is decided by the stopping criteria, where we stop querying only when the posterior saturates at a desired level. This also reflects the model's confidence in the prediction after certain number of queries. It's worth noting that the number of queries used for each sample can vary for FM+V-IP, whereas the number of concepts used is fixed for once the hyperparameters is selected.

\section{Implementation Details}
All of our experiments are implemented using Python 3.9 with PyTorch version 1.12~\citep{paszke2017automatic}. All experiments are done on one computing node with 64-core 2.10GHz Intel(R) Xeon(R) Gold 6130 CPU, eight NVIDIA GeForce RTX 2080 GPU (each with 10GB memory) and 377GB of RAM.

\textbf{Datasets.} In this paper, we used 9 datasets in total: CUB-200~\citep{koh2020concept}, CIFAR-10 and CIFAR-100~\citep{radford2021learning}, ImageNet~\citep{deng2009imagenet}, Places365~\citep{zhou2017places}, Flower-102~\citep{nilsback2008automated}, FGVC-Aircraft (Aircraft)~\citep{maji13fine-grained}, and UCF-101~\citep{soomro2012ucf101}. For CUB-200, CIFAR-10 and CIFAR-100, ImageNet, Places365, we use the same training and testing split as Lf-CBM~\citep{oikarinen2023label}. For Flower-102, FGVC-Aircraft (Aircraft), and UCF-101, we use the same training, validation, and testing split as LaBo~\citep{yang2022language}. For summary, the size of each split and the number of classes for each dataset is listed in Table~\ref{tab:split_size}.

\begin{table}[h]
\resizebox{\textwidth}{!}{
\begin{tabular}{l|cccccccc}
\toprule
Dataset & CUB-200 & CIFAR-10 & CIFAR-100 & ImageNet & Places365 & Flower-102 & Aircraft & UCF-101 \\
\midrule
\# Train& 4,796 & 50,000 & 50,000 & 1.3 mil. & 1.8 mil. & 4,093 & 3,334 & 7.639 \\
\# Validation & 1,198 & - & - & 50,000 & 36,000 & 1,633 & 3,333 & 1,898 \\
\# Test & 5,794 & 10,000 & 10,000 & - & - & 2,463 & 3,333 & 3,783 \\
\midrule
\# Classes & 200 & 10 & 100 & 1,000 & 365 & 102 & 102 & 101 \\
\bottomrule
\end{tabular}
}
\caption{Number of samples in training, validation, and testing splits, as well as the number of classes for each dataset.} \label{tab:split_size}
\end{table}

Note that there are multiple versions of Places365 dataset. In this work, we use the standard dataset with small images.

\textbf{Query Sets $Q$.} In Section~\ref{exp:suff} and Section~\ref{sec:compare-with-lfcbm}, we generate a query set by prompting GPT-3 with the following hyperparameters: \texttt{text-davinci-003} engine, 0 temperature, 500 maximum tokens, \texttt{top\textunderscore{p}=1}, 0 frequency penalty and 0.5 presence penalty. Query set $Q_{\text{filtered}}$ in Section~\ref{sec:compare-with-lfcbm} and query set in Section~\ref{sec:compare-with-labo} (here we denote as $Q_{\text{LaBo}}$) are directly obtained from GitHub repositories of Lf-CBM\footnote{\url{https://github.com/Trustworthy-ML-Lab/Label-free-CBM}} and LaBo\footnote{\url{https://github.com/YueYANG1996/LaBo}}. The size of each query set is listed in Table~\ref{tab:query_set_size}:
\begin{table}[h]
\resizebox{\textwidth}{!}{
\begin{tabular}{l|cccccccc}
\toprule
Size & CUB-200 & CIFAR-10 & CIFAR-100 & ImageNet & Places365 & Flower-102 & Aircraft & UCF-101 \\
\midrule
$Q_{\text{filtered}}$ & 208 & 128 & 824 & 4523 & 2207 & - & - & - \\
$Q_{\text{GPT}}$ & 518 & 204 & 1152 & 4495 & 2491 & - & - & - \\
$Q_{\text{LaBo}}$ & - & - & - & - & - & 24960 & 36324 & 46845 \\
\bottomrule
\end{tabular}
}
\caption{Query set sizes for different datasets. $Q_{\text{filtered}}$ was used in Section~\ref{sec:compare-with-lfcbm}, and is referenced directly from Lf-CBM~\citep{oikarinen2023label}. $Q_{\text{GPT}}$ was used in Section~\ref{exp:suff} and Section~\ref{sec:compare-with-labo}. $Q_{\text{LaBo}}$ is used in Section~\ref{sec:compare-with-labo}, and is referenced directly from LaBo~\citep{yang2022language}.}\label{tab:query_set_size}
\end{table}

\textbf{Query Answers $Q(X)$.} Query answers are computed using CLIP dot-products. Again, we denote $\mathcal{I}(\cdot)$ and $\mathcal{T}(\cdot)$ as the image and text encoder from CLIP. For any task and dataset, we obtain the image embeddings of all images $\mathcal{I}(x) \ \forall x \in \mathcal{X}$ and $\mathcal{T}(q) \ \forall q \in Q$. Each embedding is assumed to be $\ell_2$-normalized, then the query answer $q(x) = \mathcal{I}(x) \cdot \mathcal{T}(q)$. Following Lf-CBM, we further Z-score standardize each query answer by subtracting the mean and dividing the standard deviation of query answers from the training set:
\begin{equation}
    q(x) \leftarrow \frac{q(x) - \hat{\mu}}{\hat{\sigma}}, \quad \hat{\mu} = \frac{1}{|\mathcal{X}| |Q|}\sum_{x \in \mathcal{X}} \sum_{q_\in Q} q(x), \quad \hat{\sigma} = \sqrt{\sum_{x \in \mathcal{X}} \sum_{q \in Q} \frac{(q(x) - \hat{\mu})^2}{|\mathcal{X}||Q|}},
\end{equation}
which can be computed easily with \texttt{np.mean} and \texttt{np.std}.

For experiments in Section~\ref{exp:suff} and Section~\ref{sec:compare-with-lfcbm}, we use CLIP with \texttt{ViT-B/16} backbone. For experiments in Section~\ref{sec:compare-with-labo}, we use the same backbone as Labo and use CLIP with \texttt{ViT-L/14} backbone ~\citep{yang2022language}. Each image is preprocessed with CLIP's default \texttt{preprocess} function ~\citep{radford2021learning}.

\textbf{Representing and Updating the History $S$.}
In V-IP framework, the input to predictor $f_\theta$ and querier $g_\eta$ are both histories $S$. Similar to~\citet{chattopadhyay2023variational}, the history for a sample $x$, $S(x)$, is represented as the product of $|Q|$-dimensional feature vector for all query answers $Q(x)$ and a binary mask $M$, i.e. $Q(x) \odot M$, where $\odot$ here represents point-wise multiplication, also known as the Hadamard product. The $i$-th dimension corresponds to the query answer $q_i(x)$. We set $M_i$ to $0$ if the $q_i(x) \not \in S$ and $1$ if $q_i(x)  \in S$. Hence, a history with all query answers has $M$ equal to a vector of all ones, and an empty history is represented by a mask of all zeros.

Suppose we have a history for sample $x$ of size $k$, $S_k(x) = Q(x) \cdot M_k$. To update the history $S_k$ with an additional query from the output of the querier $g_\eta(S_k)$, we simply update the mask and the history by:
\begin{equation*}
    M_{k+1} \leftarrow M_k + g_\eta(S_k) \quad \text{and} \quad S_{k+1} \leftarrow S_k + Q(x) \odot g_\eta(S_k)
\end{equation*}
In other words, the new query is mathematically represented as a one-hot vector with dimension $|Q|$, hence updating the representation of history is equivalent to updating the binary mask $M$ by setting $i$-th position to 1, indicating that the $q_i(x)$ is in the updated history $S_{k+1}(x)$.

\begin{figure}[ht]
    \centering
    \begin{subfigure}{0.99\textwidth}
        \centering
        \includegraphics[width=0.8\textwidth]{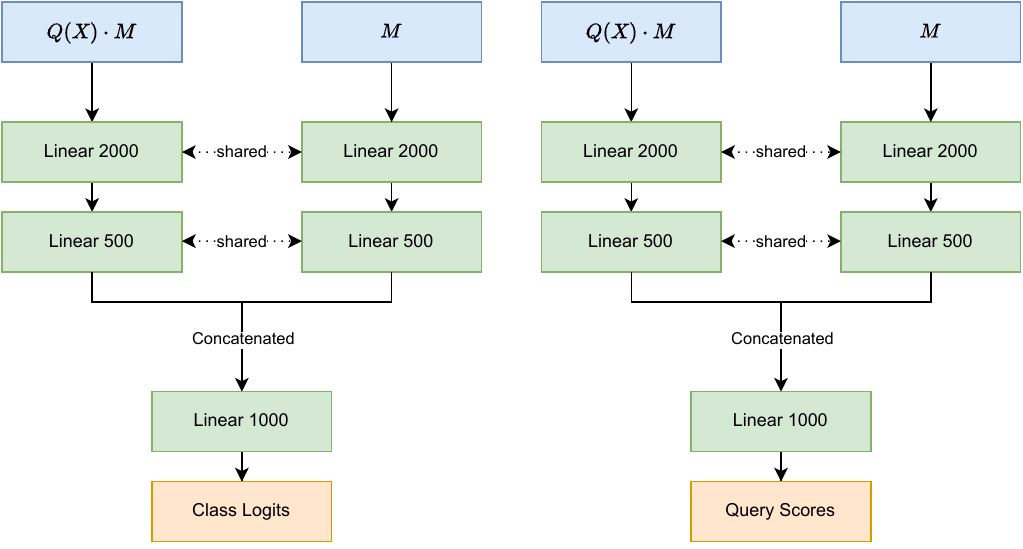}
        \caption{\texttt{shallow} architecture for \textbf{(left)} predictor $f_\theta$ and \textbf{(right)} querier $g_\eta$.}
        \vspace{0.5in}
    \end{subfigure}
    
    \begin{subfigure}{0.99\textwidth}
        \centering
        \includegraphics[width=0.8\textwidth]{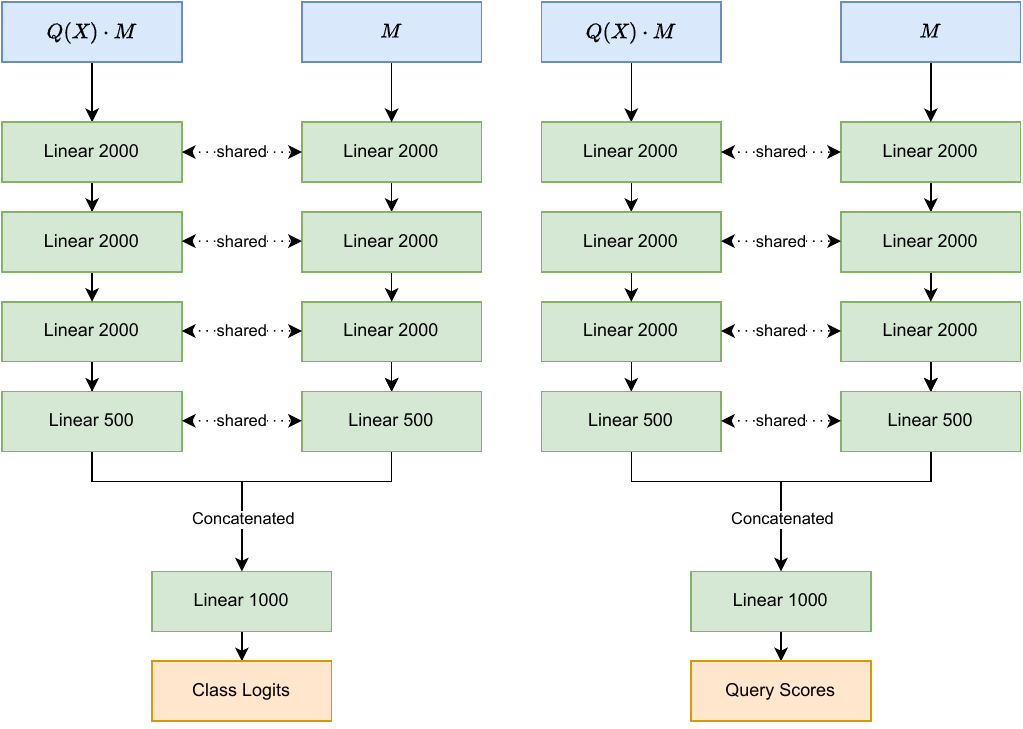}
        \caption{\texttt{deep} architecture for \textbf{(left)} predictor $f_\theta$ and \textbf{(right)} querier $g_\eta$.}
    \end{subfigure}
    
    \caption{Architecture designs for \texttt{shallow} and \texttt{deep}. ``shared'' implies the weights are shared between the two linear layers. ``Concatenated'' implies the output from previous layers are concatenated ($a \in \mathbb{R}^{1 \times n}, b \in \mathbb{R}^{1 \times }, \mathrm{concat(a, b) = [a | b] \in \mathbb{R}^{1 \times (n+m)}}$). Every arrow $\rightarrow$ before the concatenation and after the input layer is LayerNorm of appropriate dimension, followed by ReLU.}\label{app:arch}
\end{figure}

\textbf{Architectures.} In this work, we only have two design choices for the predictor $f_\theta$ and querier $g_\eta$, which we will denote them as \texttt{shallow} and \texttt{deep}. \texttt{shallow} is a two-layer fully connected neural network, and \texttt{deep} is a four-layer fully connected neural network. Diagrams of their architecture is shown in Figure~\ref{app:arch}. The \texttt{shallow} architecture is used for medium-scale datasets CIFAR-10, CIFAR-100, CUB-200, Flower-102, FVGC-Aircraft, and UCF-101. The \texttt{deep} architecture is used for large-scale datasets ImageNet and Places365. {The number of parameters for each experiment is listed in Table~\ref{tab:n-params}. The architecture design is chosen empirically based on the performance of the model. The size of the query set only affects the final output dimension of the querier $g_\eta$, and the number of classes affects the final output dimension of the classifier $f_\theta$. }In all of our experiments, we do not share the weights between $f_\theta$ and $g_\eta$. We apply Softmax operator to class logits and query scores to obtain probability for each class and probability for each query.During training, a Straight-through softmax with temperature parameter $\tau$ is used for computing query probabilities. For every experiment, we linear decay $\tau$ from $1.0$ to $0.2$ for 20 epochs \footnote{$\tau = 1.0$ is equivalent to regular Softmax operator, whereas $\tau \rightarrow 0$ corresponds to $\mathrm{argmax}(\cdot)$ operator.}. We find V-IP training is rather insensitive to how we anneal $\tau$. 

\begin{table}[h]
\resizebox{\textwidth}{!}{
\begin{tabular}{lcccccccc}
\toprule
\# of Parameters      & CUB-200 & CIFAR-10 & CIFAR-100 & ImageNet & Places365 & Flower-102 & Aircraft  & UCF-101   \\
\midrule
$Q_{\text{filtered}}$ & 3,051,408 & 2,596,138  & 5,773,924   & 41,659,523 & 29,441,572  & -          & -         & -         \\
$Q_{\text{GPT}}$      & 4,446,718 & 2,938,214  & 7,250,252   & 41,519,495 & 3,0861,856  & -          & -         & -         \\
$Q_{\text{LaBo}}$     & -       & -        & -         & -        & -         & 126,942,062  & 183,773,426 & 236,387,946 \\
\bottomrule
\end{tabular}
\caption{The total number of parameters for each dataset.}\label{tab:n-params}
}
\end{table}

\textbf{Training FM+V-IP.} Every FM+V-IP experiment for medium-scale datasets CIFAR-10, CIFAR-100, CUB-200, Flower-102, Aircraft, and UCF-101 follow the two-stage training procedure mentioned in the \citet{chattopadhyay2023variational}, where we optimize the V-IP objective for $f_\theta$ and $g_\eta$ with random histories, by first using the Randomly Sampling strategy for $4000$ epochs with learning rate $0.0001$, followed by the Subsequent Biased Sampling strategy for $1500$ epochs with learning rate $0.00005$. In both stages, we train using Adam Optimizer~\citep{kingma2014adam} with no weight decay, along with Cosine Annealing learning rate scheduler and hyperparameter \texttt{T\textunderscore{max}=100}. 

Similarly, for large-scale datasets ImageNet and Places365, we also follow the two-stage training procedure. We use the same optimizer and learning rate scheduler with the same optimizer hyperparameter as mentioned above, but we train the Random Sampling stage for $400$ epochs, and train the Subsequent Biased Sampling stage for $40$ epochs. In both stages, we also train using Adam Optimizer~\citep{kingma2014adam} with no weight decay, but with Cosine Annealing learning rate scheduler and hyperparameter \texttt{T\textunderscore{max}=100}.

\textbf{Training Baselines.} In Figure~\ref{fig:vip_vs_lfcbm}, we compare the test performance of FM+V-IP with Lf-CBM for different query sets, along with with baselines, where we train the predictor for V-IP without the querier on all query answers. For fair comparison, classifier for each dataset is the same as the predictor from training FM+V-IP. For medium-scale datasets, we train for $1000$ epochs using Adam Optimizer with learning rate $0.0001$ and no weight decay, and Cosine Annealing learning rate scheduler with hyperparameter \texttt{T\textunderscore{max}=100}.

\textbf{Training Lf-CBMs.} In Section~\ref{sec:compare-with-lfcbm}, we trained multiple Lf-CBMs with varying sparsity level $\lambda$ using the unfiltered query set $Q_{\text{GPT}}$. To train these models, we optimize using GLM-SAGA optimizer~\citep{defazio2014saga} with the same hyperparameters as mentioned in Lf-CBM~\citep{oikarinen2023label}. We obtain results with different sparsity levels with $\lambda \in \{0.05, 0.045, 0.0401, 0.03516, 0.03022, 0.02527, 0.02033, 0.01538, 0.01044, 0.0055\}$ for CUB-200 and with $\lambda \in \{0.101, 0.0101, 0.005051, 0.002121, 0.001414\}$ for CIFAR-10.

{ 
\textbf{Inference Time.} Inference on one sample involves performing multiple forward pass on the querier $g$ to obtain the next optimal query and the classifier $f$ to obtain the posterior distribution $P(Y | S)$. Using our hardware, one forward pass takes no more than 0.1 seconds. Inference speed can be further improved by batch processing on multiple samples, depending on hardware memory limitations. For each sample, we begin with an empty set, and append new queries selected by the querier. In contrast, inference on Lf-CBM, LaBo or vanilla NNs are faster, since they only involve a single forward pass, either on a single linear map or on a nonlinear neural network. Nevertheless, relatively speaking, we believe believe a 0.1 second overhead in inference time to be minuscule.

}
\section{Ablation Study}
\subsection{CLIP Text Embeddings: Prompt-tuning for Queries}
\begin{figure}[h]
    \centering
    \includegraphics[width=0.5\textwidth]{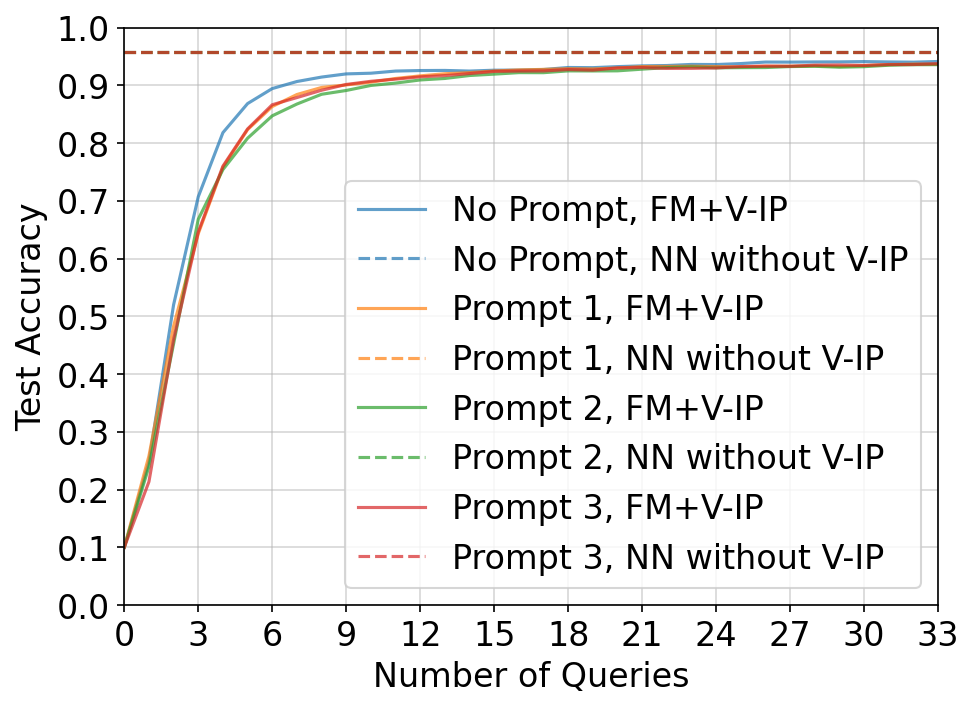}
    \caption{Test performance of FM+V-IP using different prompts for compute text embeddings for queries. ``FM+VIP'' are results trained with V-IP,  where as ``NN without V-IP'' are results from training the classifier only with all query answers. }
    \label{fig:app_abl_clip-prompts}
\end{figure}
\citet{radford2021learning} has demonstrated in their work that CLIP can achieve state-of-the-art test performance in zero-shot image classification tasks. To do so, they first compute the image embeddings for every test image. Then for each class name \texttt{\{class\textunderscore{name}\}}, they use multiple possible prompts such as ``a bright photo of a \texttt{\{class\textunderscore{name}\}},'' ``a photo of a dirty \texttt{\{class\textunderscore{name}\}},'' ``a photo of a small \texttt{\{class\textunderscore{name}\}}.''
Intuitively, the optimal prompt should predict the correct class with a higher score (dot-product between image and text embeddings). Here, we perform a similar study, where we compare the difference in test performance when using four different prompts given a query/concept \texttt{\{concept\}}:
\begin{enumerate}
    \item \textit{No Prompt}: \texttt{\{concept\}}
    \item \textit{Prompt 1}: An image of an object with \texttt{\{concept\}}.
    \item \textit{Prompt 2}: An image of an small object with \texttt{\{concept\}}.
    \item \textit{Prompt 3}: A small image of an object with \texttt{\{concept\}}.
\end{enumerate}
Our results for CIFAR-10 are shown in Figure~\ref{fig:app_abl_clip-prompts}. We find that using the different prompts result in little difference in test performance for FM+V-IP. From our baselines, which are models from training the classifier only with all query answers, we also see there is little-to-no difference in test performance.

\subsection{CLIP Image Embeddings: Using Different Backbones}

\begin{figure}[h]
    \centering
    \includegraphics[width=0.5\textwidth]{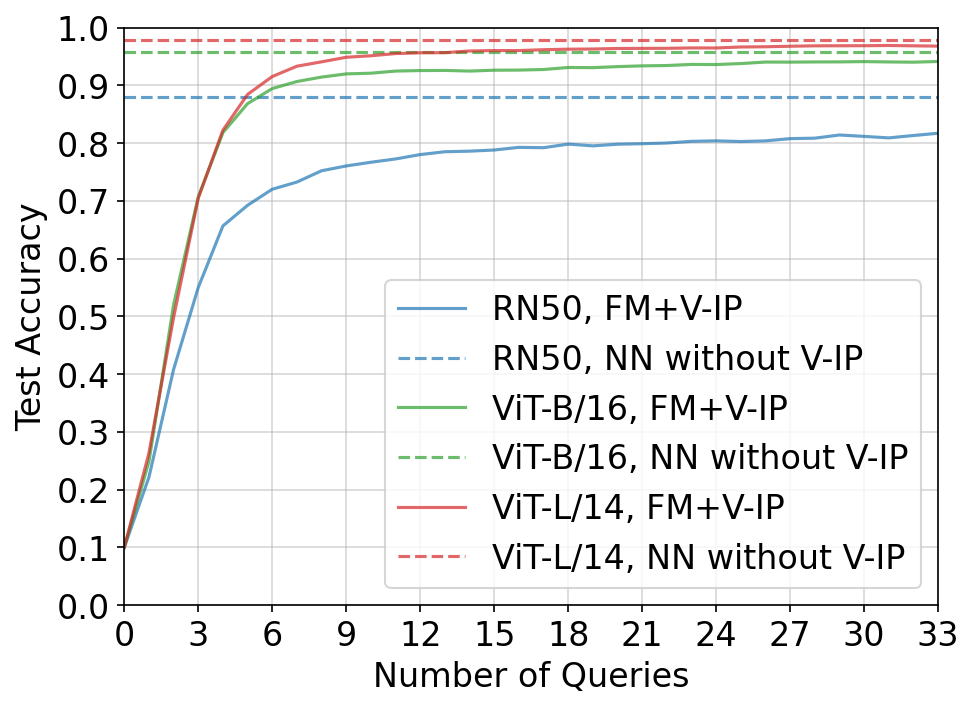}
    \caption{Test performance of FM+V-IP with different backbones. }
    \label{fig:app_abl_backbone}
\end{figure}
As mentioned above, CLIP provides the choice of choosing a backbone as the image encoder. Reasonably, the more parameters the backbone has, the better the quality of the text embeddings should be. Here, we experiment with the following three different backbones and observe the impact the choice of backbone has on test performance for FM+V-IP:
\begin{enumerate}
    \item \texttt{RN50}:  ResNet-50. This architecture has the least parameters. 
    \item \texttt{ViT-B\textbackslash16}: Vision Transformer with Base architecture with $16\times 16$ image patches
    \item \texttt{ViT-L\textbackslash14}: Vision Transformer with Large architecture with $14 \times 14 $ image patches. This architecture has the most parameters. 
\end{enumerate}

Our results for CIFAR-10 are shown in Figure~\ref{fig:app_abl_backbone}. As expected, the encoder with more parameters have higher quality image embeddings, the query answers are less noisy and help FM+V-IP achieves better performance. Notably, at 15 queries, FM+V-IP using \texttt{ViT-L\textbackslash14} achieves a 96\% test accuracy, whereas \texttt{RN50} achieves a 79\% test accuracy, a difference of about 17\%. From this, we can conclude that the number of parameters in the image encoder backbone has a big impact on the test performance of FM+V-IP.

    

\section{Further Examples of Query-Answer Chains}
\begin{figure}
    \centering
    \begin{subfigure}[b]{0.7\textwidth}
        \includegraphics[width=\textwidth]{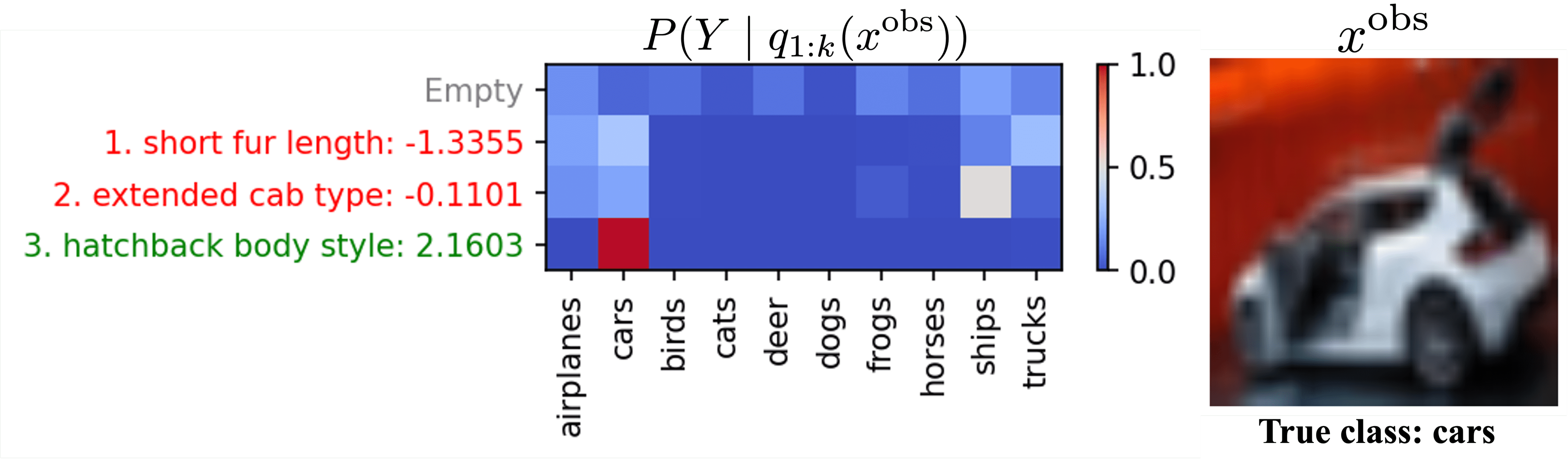}
    \end{subfigure}
    \begin{subfigure}[b]{0.7\textwidth}
        \includegraphics[width=\textwidth]{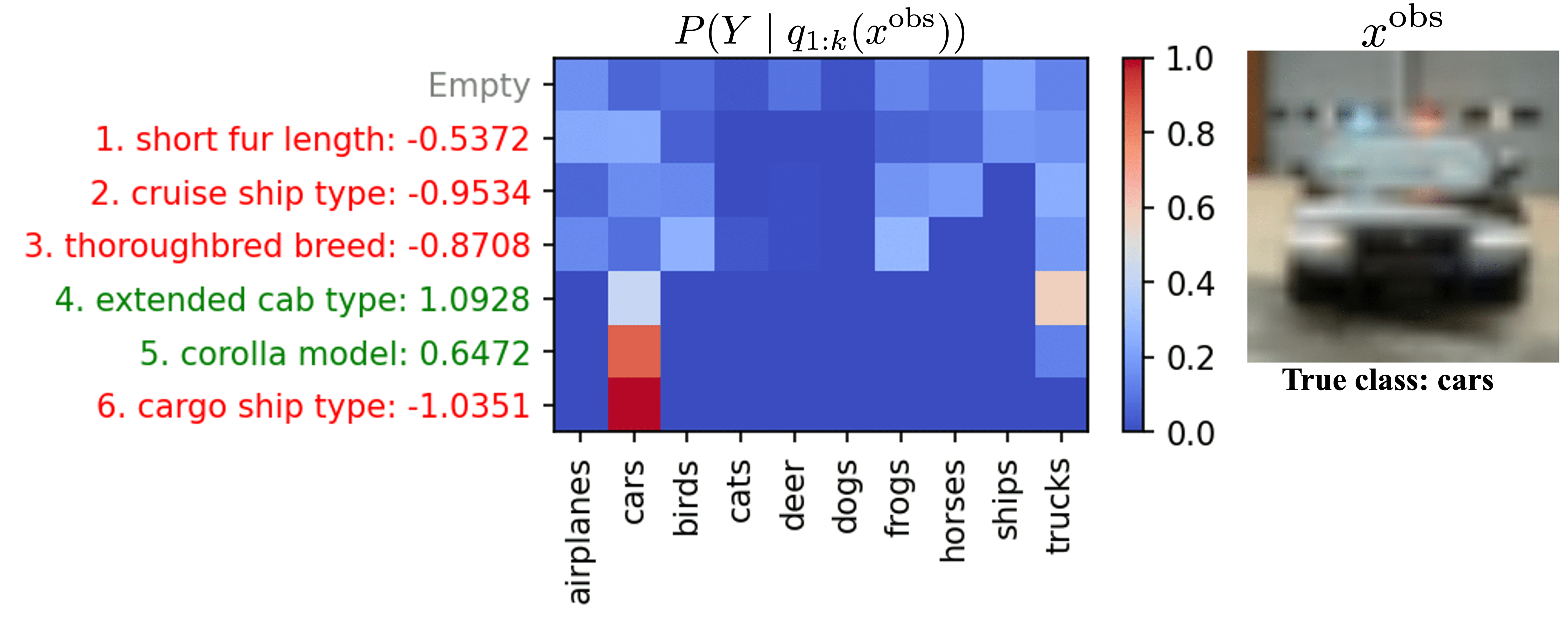}
    \end{subfigure}
    \begin{subfigure}[b]{0.7\textwidth}
        \includegraphics[width=\textwidth]{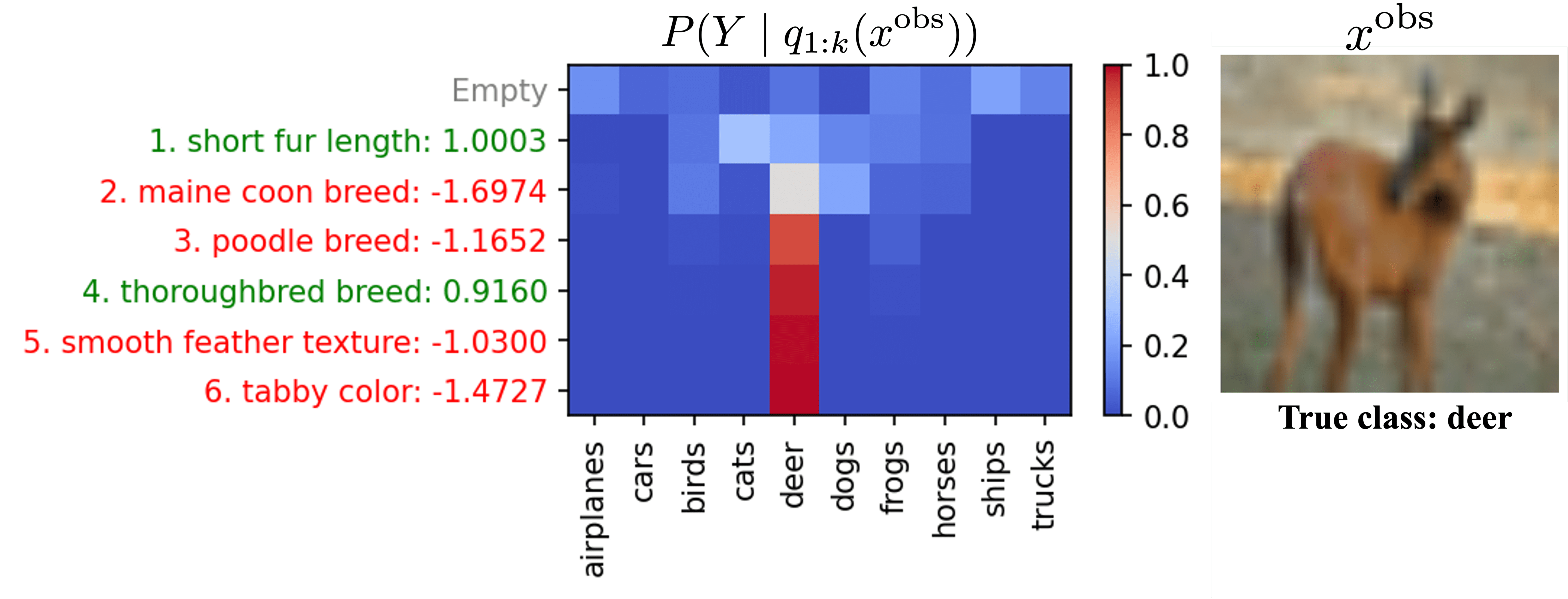}
    \end{subfigure}
    \begin{subfigure}[b]{0.7\textwidth}
        \includegraphics[width=\textwidth]{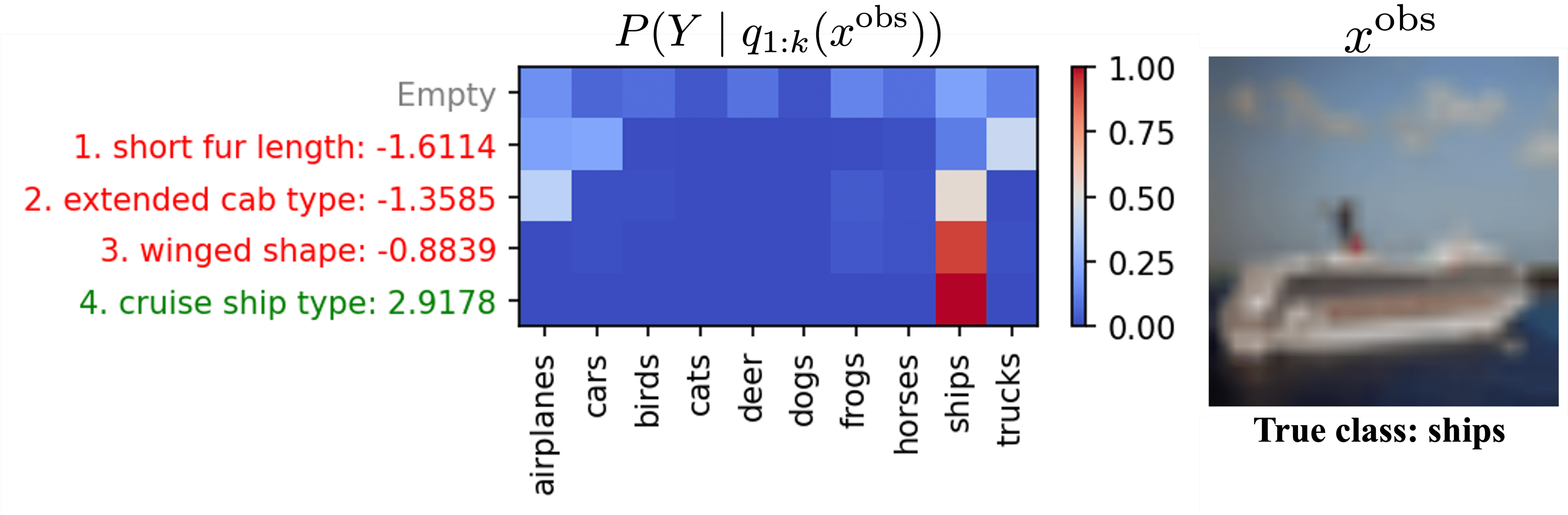}
    \end{subfigure}
    \begin{subfigure}[b]{0.7\textwidth}
        \includegraphics[width=\textwidth]{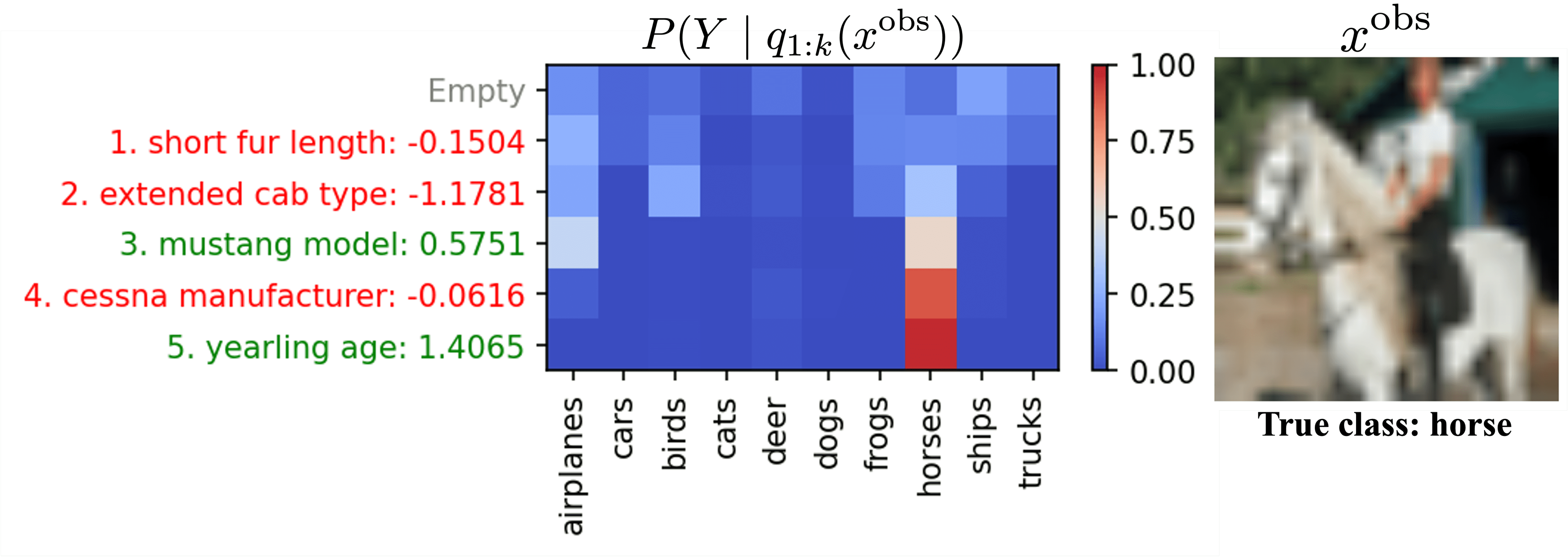}
    \end{subfigure}
    \caption{Five examples of query-answer chains from CIFAR-10.}\label{app:traj-cifar10}
\end{figure}

\begin{figure}
    \centering
    \begin{subfigure}[b]{0.7\textwidth}
        \includegraphics[width=\textwidth]{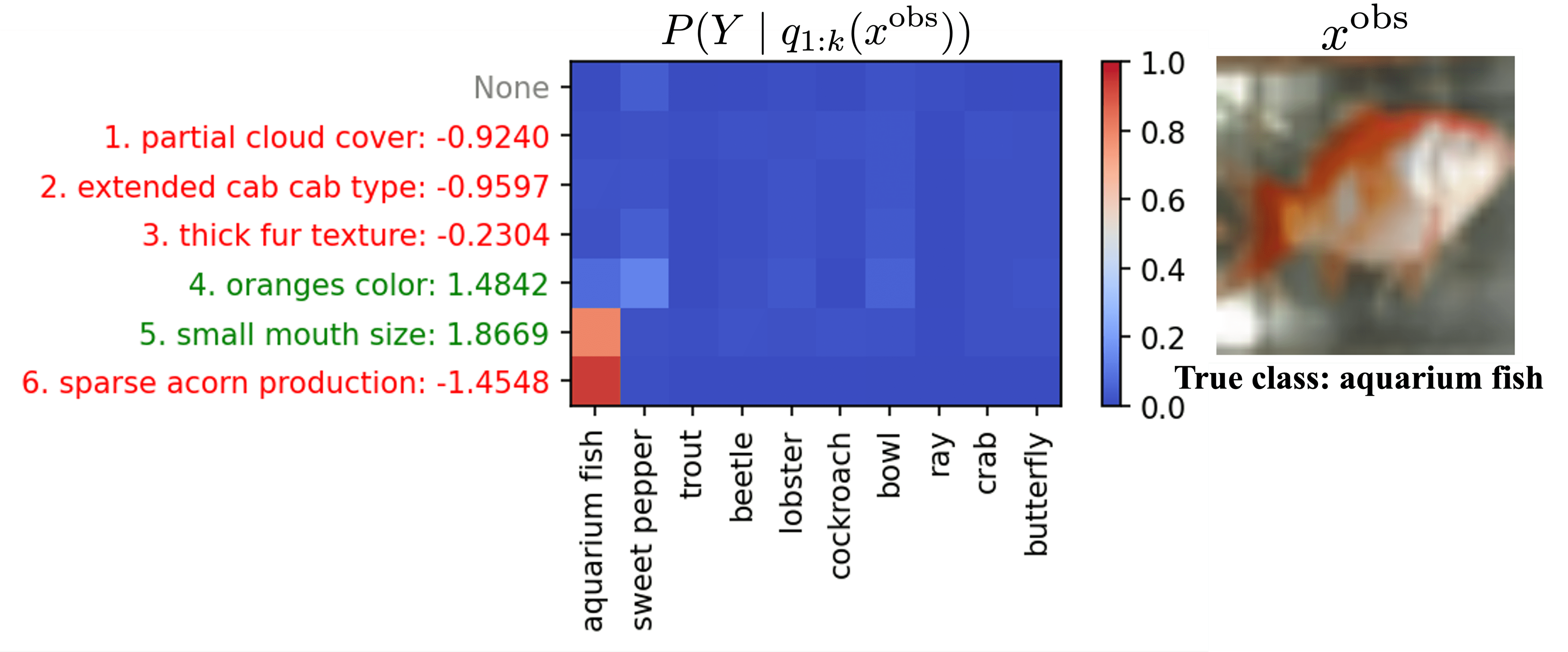}
    \end{subfigure}
    \begin{subfigure}[b]{0.7\textwidth}
        \includegraphics[width=\textwidth]{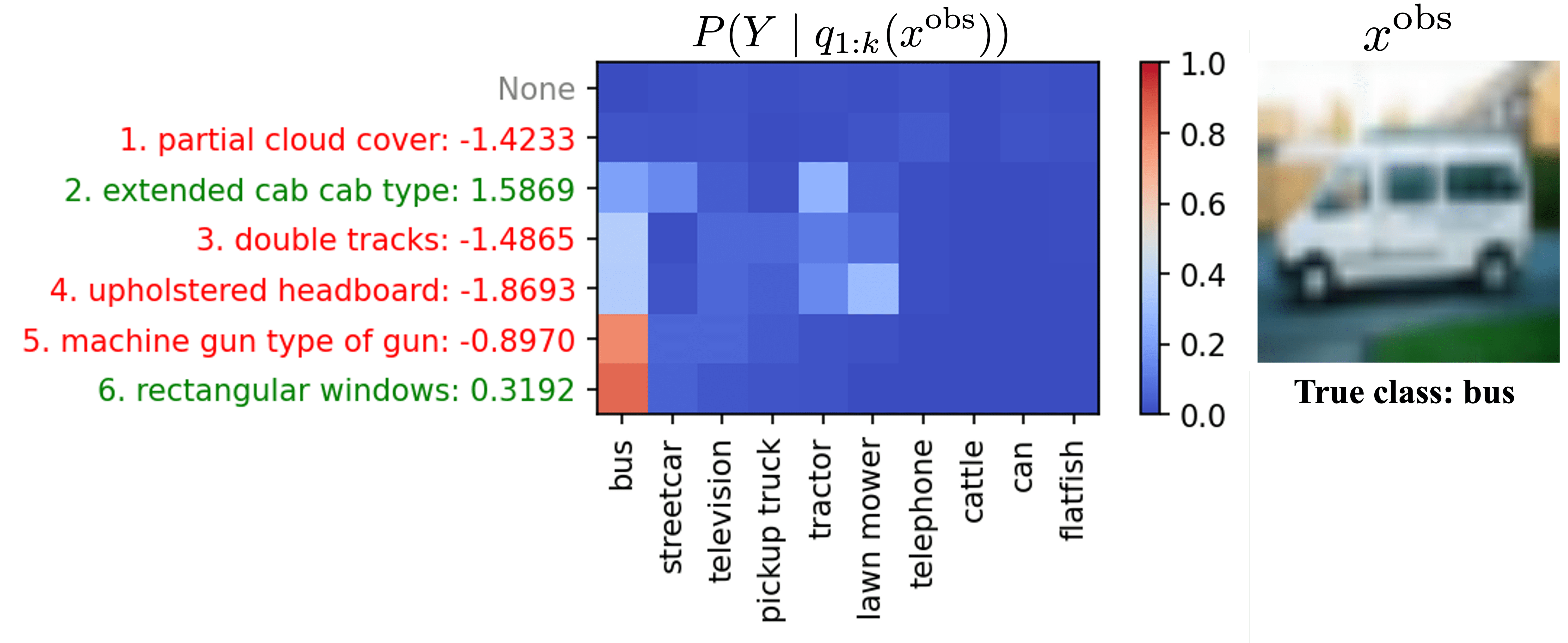}
    \end{subfigure}
    \begin{subfigure}[b]{0.7\textwidth}
        \includegraphics[width=\textwidth]{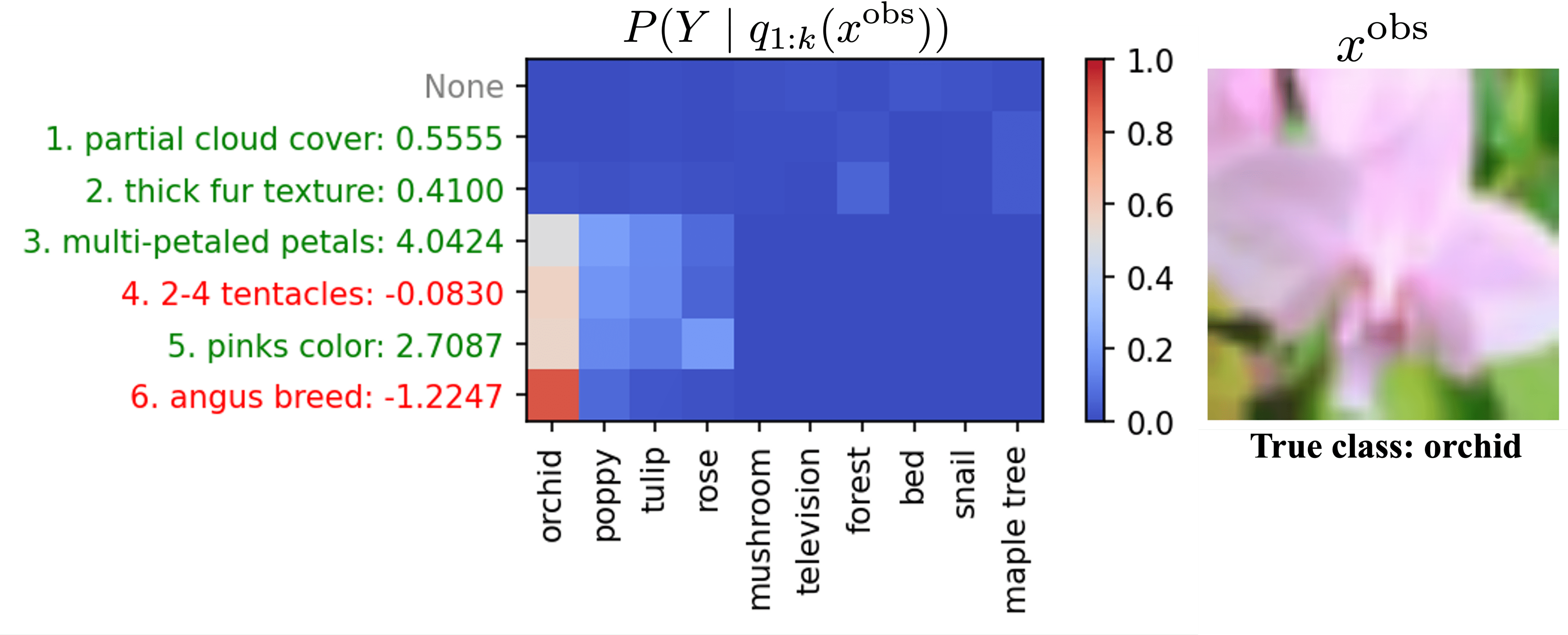}
    \end{subfigure}
    \begin{subfigure}[b]{0.7\textwidth}
        \includegraphics[width=\textwidth]{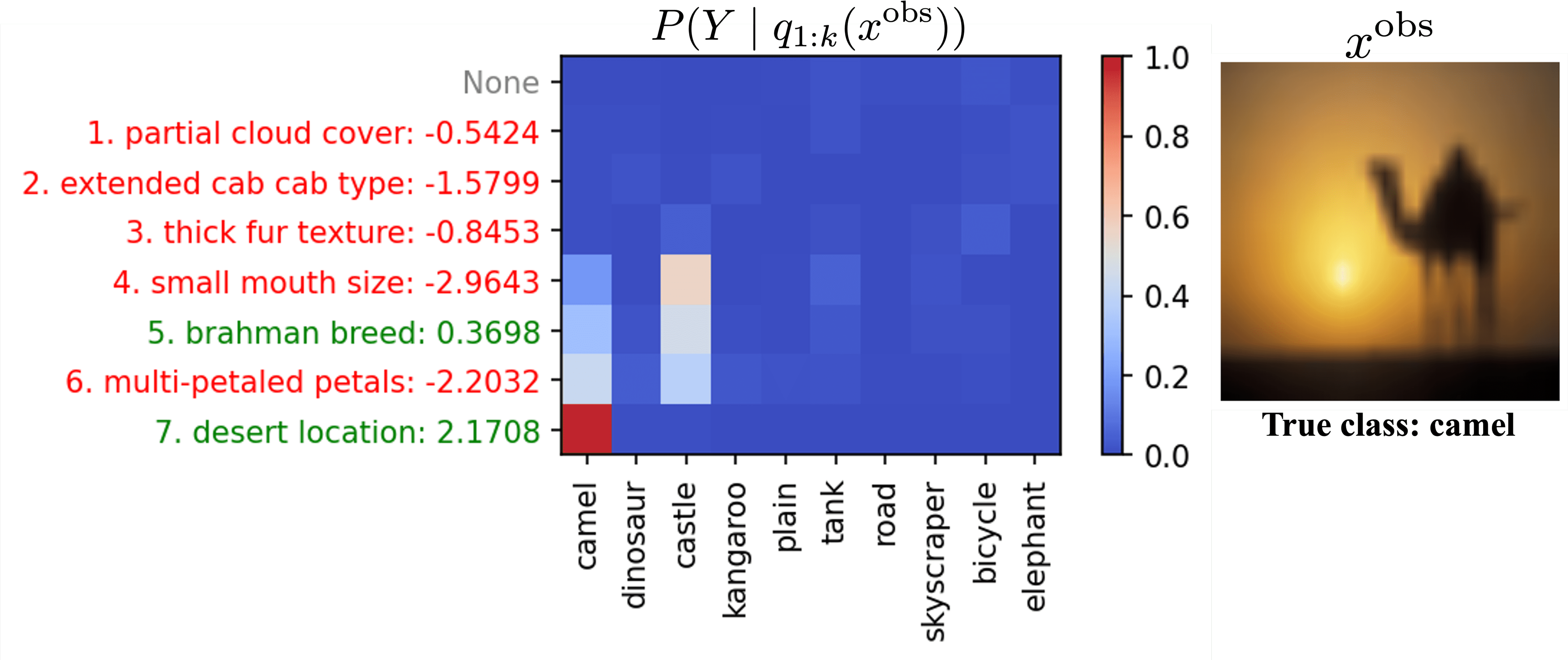}
    \end{subfigure}
    \begin{subfigure}[b]{0.7\textwidth}
        \includegraphics[width=\textwidth]{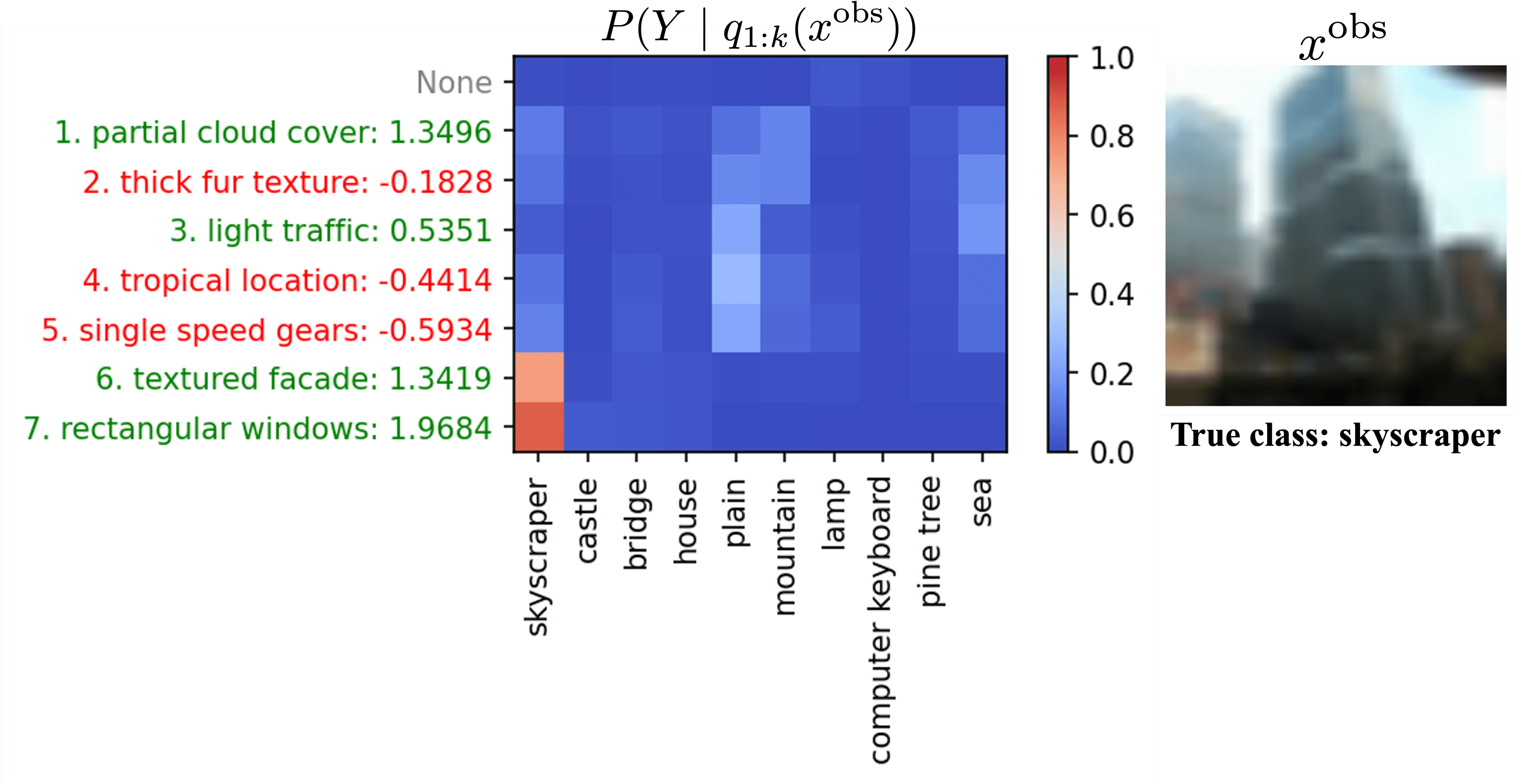}
    \end{subfigure}
    \caption{Five examples of query-answer chains from CIFAR-100.}\label{app:traj-cifar100}
\end{figure}

\begin{figure}
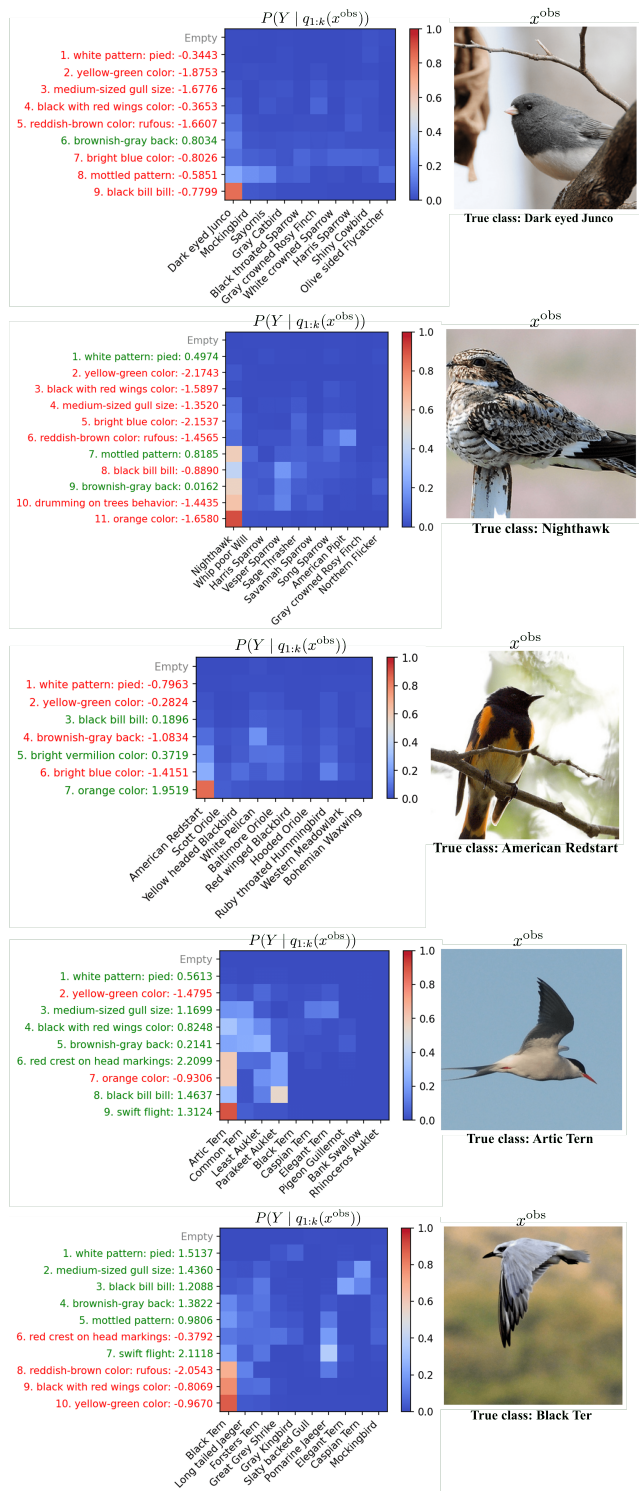

    \centering
    \begin{subfigure}[b]{0.6\textwidth}
        \includegraphics[width=\textwidth]{figures/app-trajs/cub/traj_cub_1.pdf}
    \end{subfigure}
    \begin{subfigure}[b]{0.6\textwidth}
        \includegraphics[width=\textwidth]{figures/app-trajs/cub/traj_cub_2.pdf}
    \end{subfigure}
    \begin{subfigure}[b]{0.6\textwidth}
        \includegraphics[width=\textwidth]{figures/app-trajs/cub/traj_cub_3.pdf}
    \end{subfigure}
    \begin{subfigure}[b]{0.6\textwidth}
        \includegraphics[width=\textwidth]{figures/app-trajs/cub/traj_cub_4.pdf}
    \end{subfigure}
    \begin{subfigure}[b]{0.6\textwidth}
        \includegraphics[width=\textwidth]{figures/app-trajs/cub/traj_cub_5.pdf}
    \end{subfigure}
    \caption{Five examples of query-answer chains from CUB-200.}\label{app:traj-cub}
\end{figure}

\begin{figure}
    \centering
    \begin{subfigure}[b]{0.6\textwidth}
        \includegraphics[width=\textwidth]{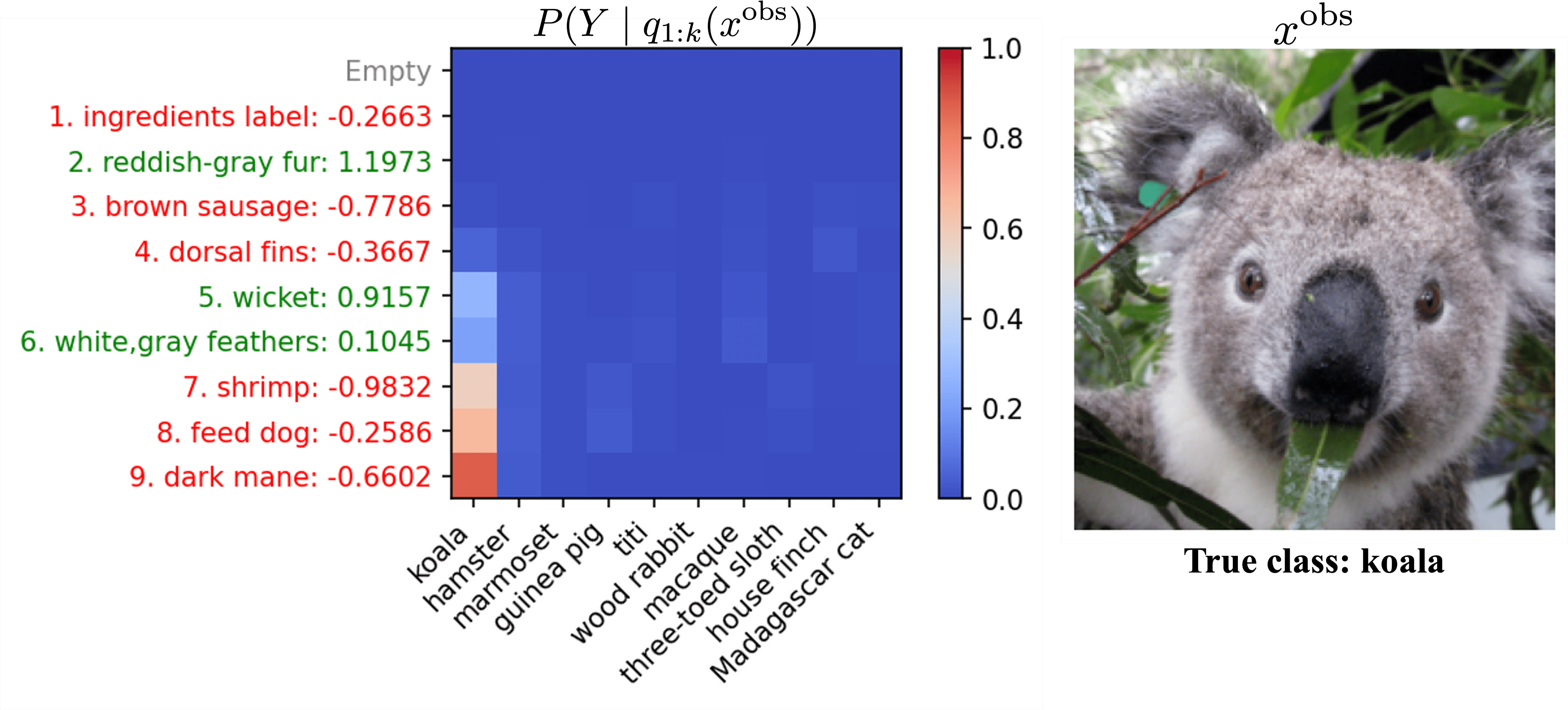}
    \end{subfigure}
    \begin{subfigure}[b]{0.6\textwidth}
        \includegraphics[width=\textwidth]{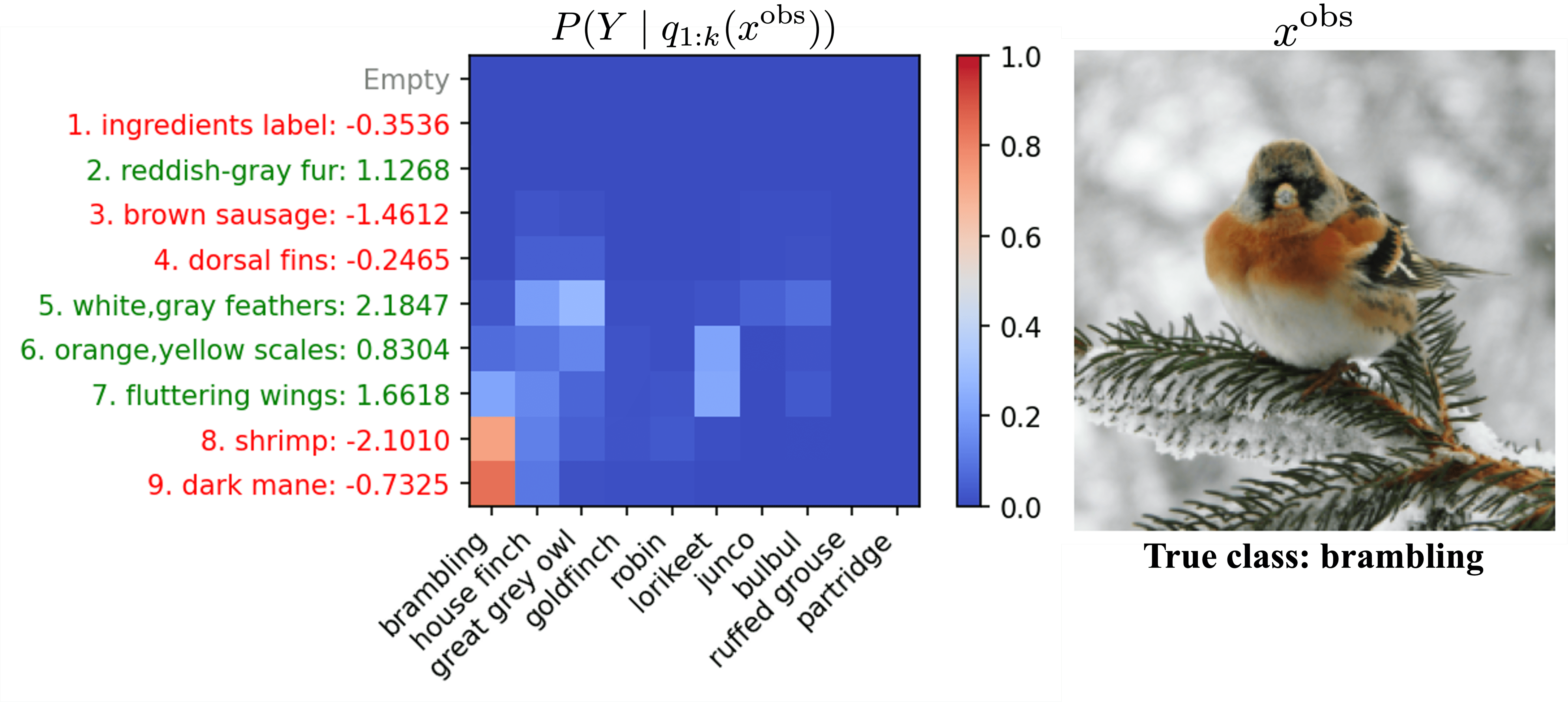}
    \end{subfigure}
    \begin{subfigure}[b]{0.6\textwidth}
        \includegraphics[width=\textwidth]{figures/app-trajs/imagenet/traj_imagenet_3.pdf}
    \end{subfigure}
    \begin{subfigure}[b]{0.6\textwidth}
        \includegraphics[width=\textwidth]{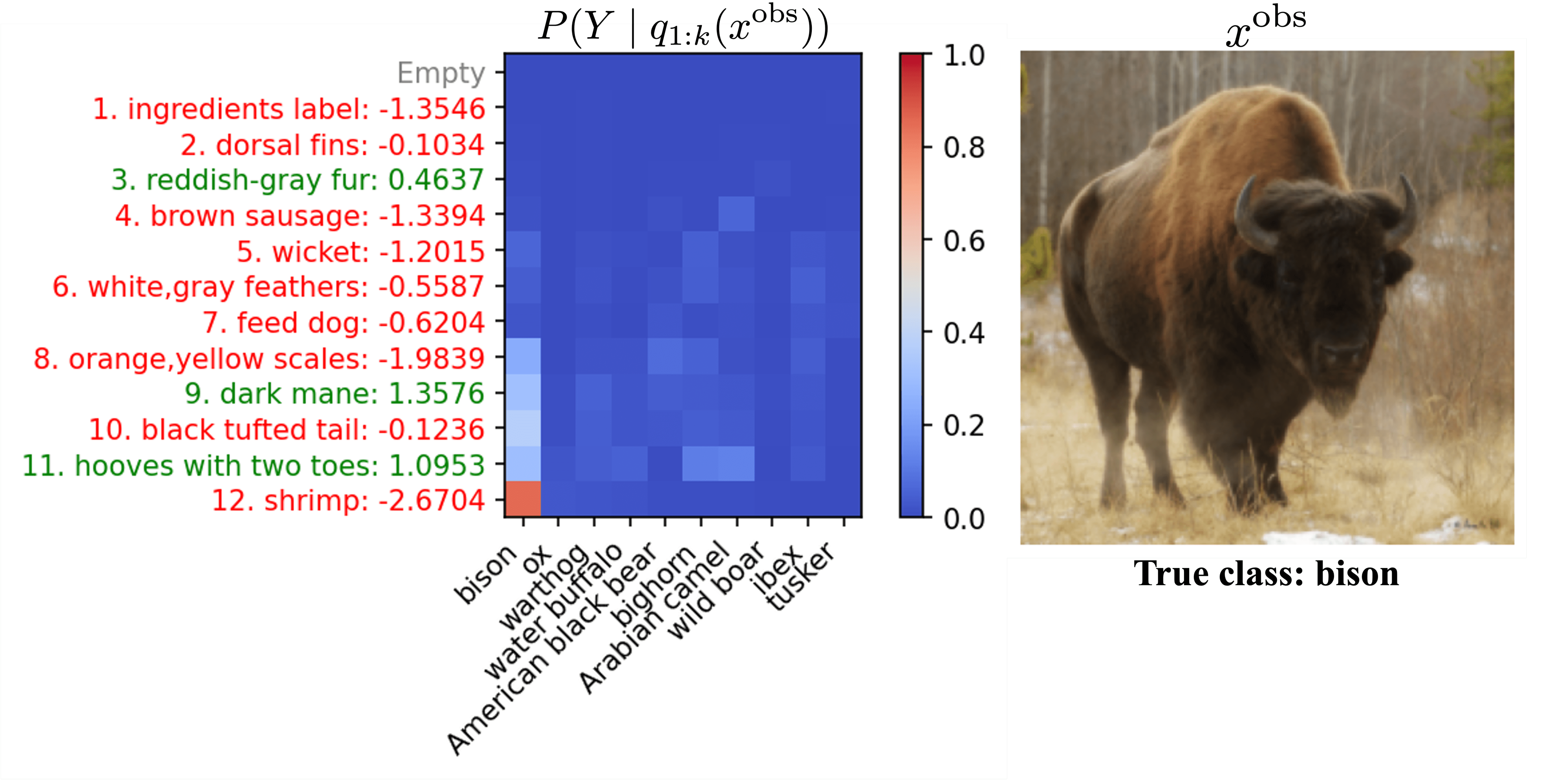}
    \end{subfigure}
    \begin{subfigure}[b]{0.6\textwidth}
        \includegraphics[width=\textwidth]{figures/app-trajs/imagenet/traj_imagenet_5.pdf}
    \end{subfigure}
    \caption{Five examples of query-answer chains from ImageNet.}\label{app:traj-imagenet}
\end{figure}

\begin{figure}
    \centering
    \begin{subfigure}[b]{0.65\textwidth}
        \includegraphics[width=\textwidth]{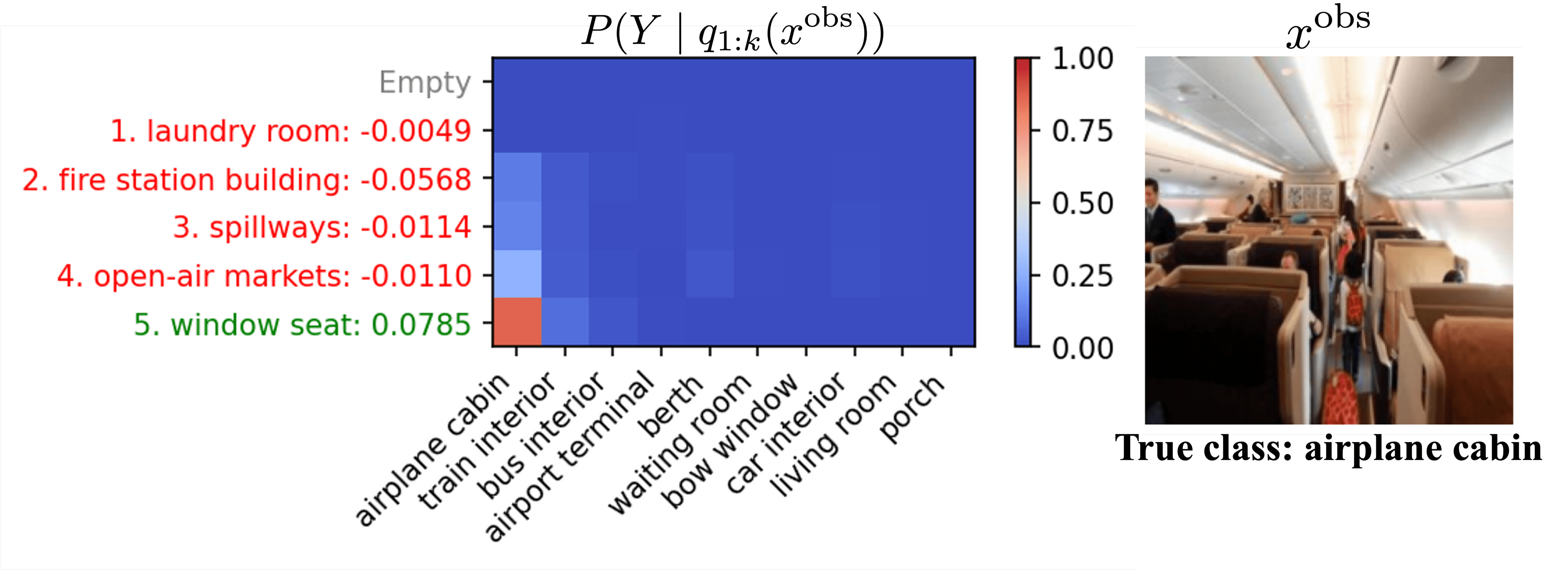}
    \end{subfigure}
    \begin{subfigure}[b]{0.65\textwidth}
        \includegraphics[width=\textwidth]{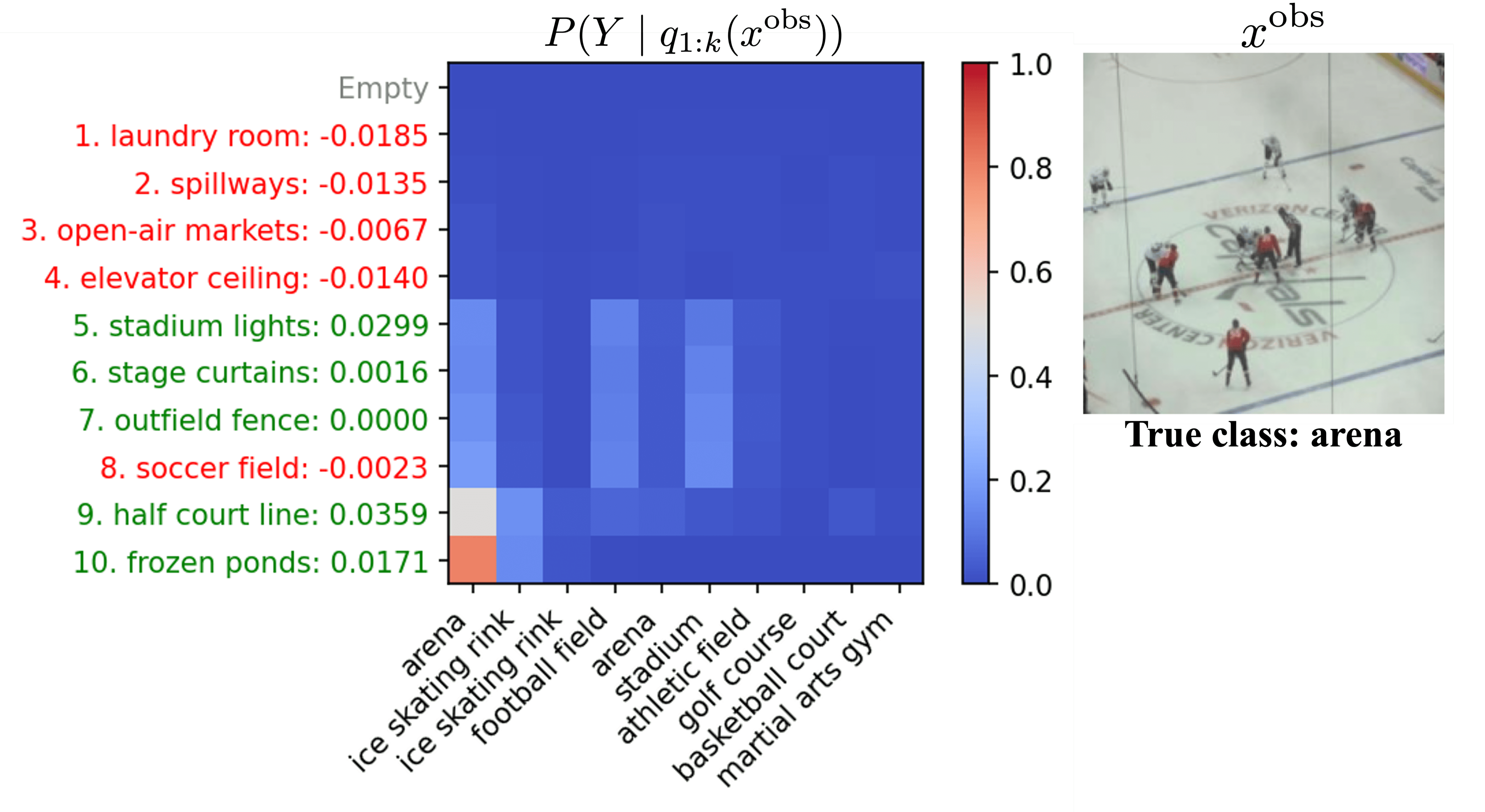}
    \end{subfigure}
    \begin{subfigure}[b]{0.65\textwidth}
        \includegraphics[width=\textwidth]{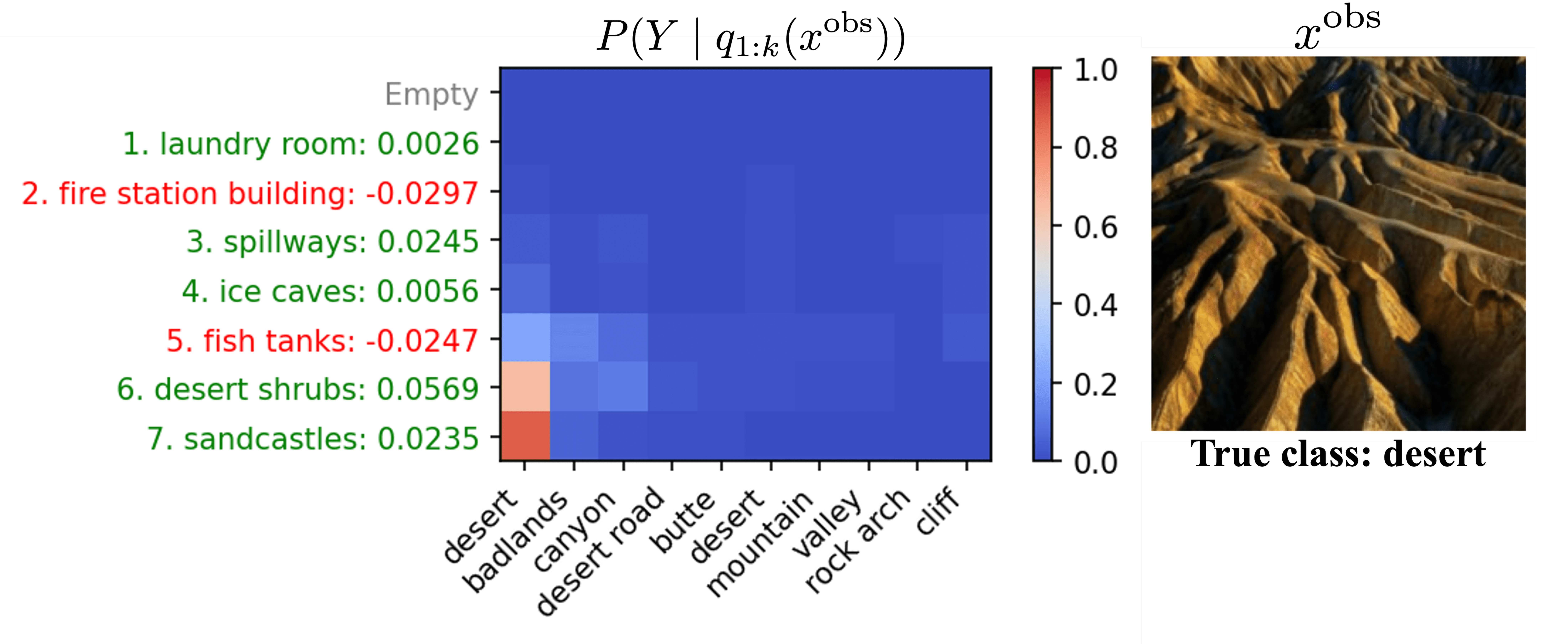}
    \end{subfigure}
    \begin{subfigure}[b]{0.65\textwidth}
    \includegraphics[width=\textwidth]{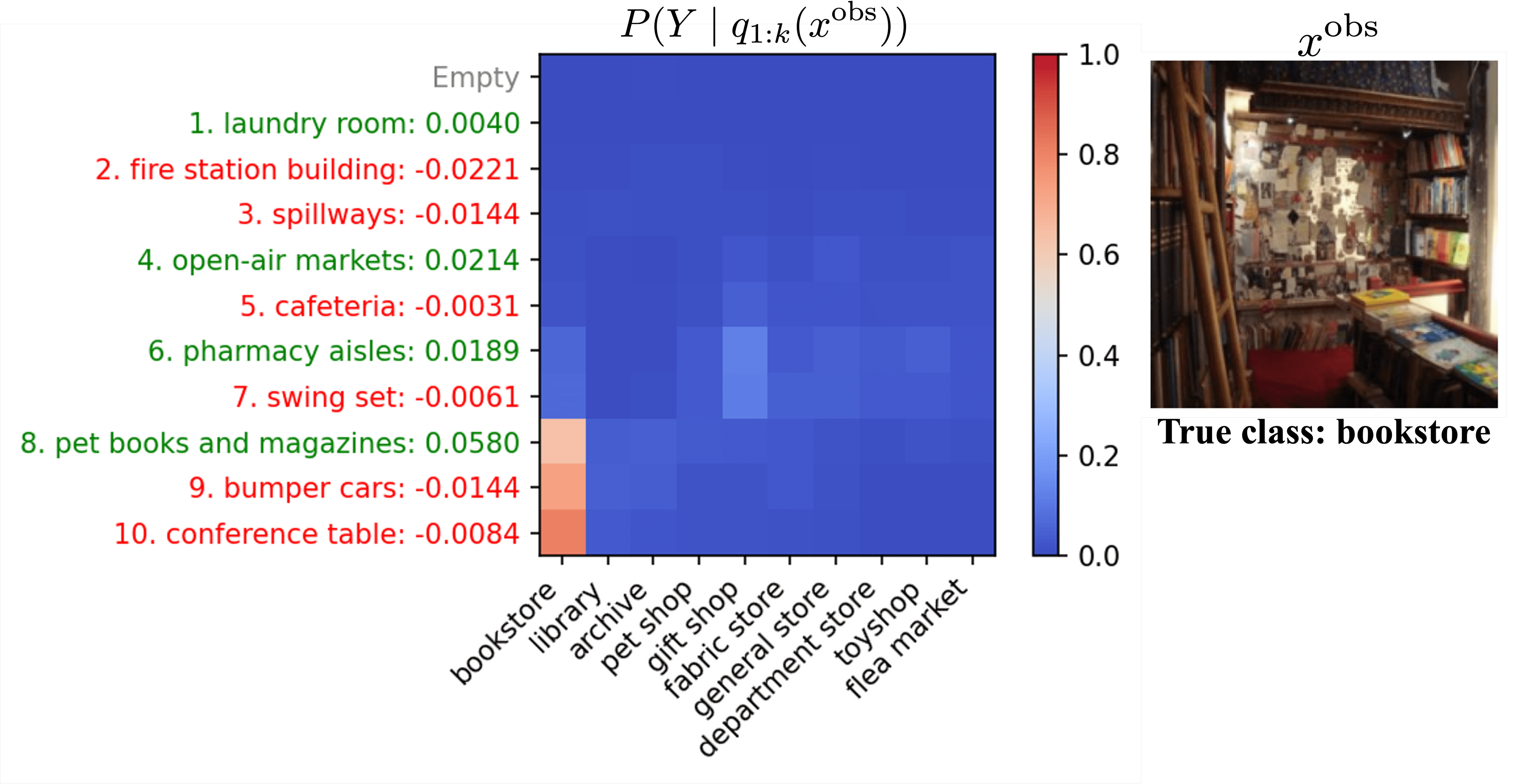}
    \end{subfigure}
    \begin{subfigure}[b]{0.65\textwidth}
    \includegraphics[width=\textwidth]{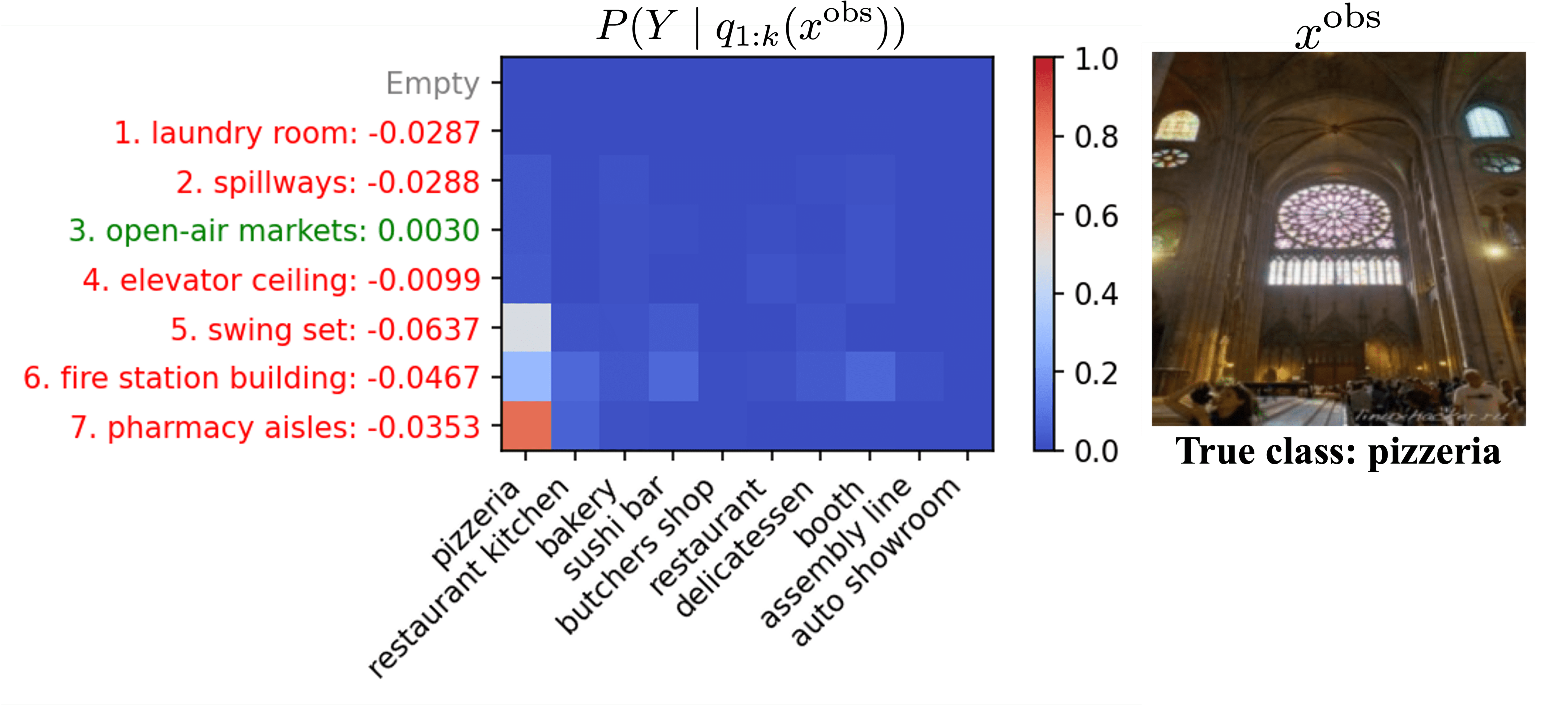}
    \end{subfigure}
    \caption{Five examples of query-answer chains from Places365.}\label{app:traj-places365}
\end{figure}

\begin{figure}
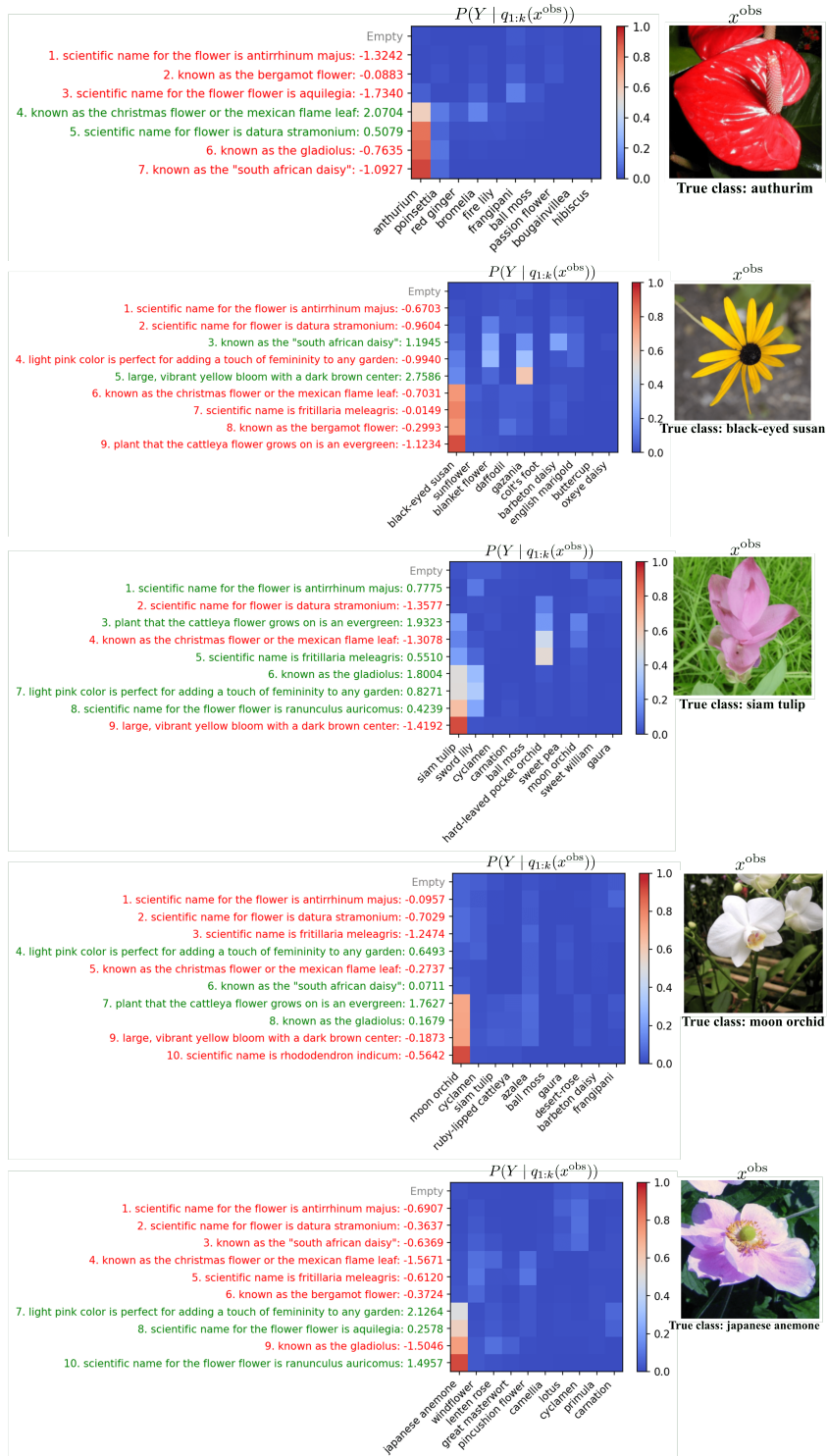

    \centering
    \begin{subfigure}[b]{0.8\textwidth}
        \includegraphics[width=\textwidth]{figures/app-trajs/flower/traj_flower_1.pdf}
    \end{subfigure}
    \begin{subfigure}[b]{0.8\textwidth}
        \includegraphics[width=\textwidth]{figures/app-trajs/flower/traj_flower_2.pdf}
    \end{subfigure}
    \begin{subfigure}[b]{0.8\textwidth}
        \includegraphics[width=\textwidth]{figures/app-trajs/flower/traj_flower_3.pdf}
    \end{subfigure}
    \begin{subfigure}[b]{0.8\textwidth}
        \includegraphics[width=\textwidth]{figures/app-trajs/flower/traj_flower_4.pdf}
    \end{subfigure}
    \begin{subfigure}[b]{0.8\textwidth}
        \includegraphics[width=\textwidth]{figures/app-trajs/flower/traj_flower_5.pdf}
    \end{subfigure}
    \caption{Five examples of query-answer chains from Flower-102.}\label{app:traj-flower}
\end{figure}

\begin{figure}
    \centering
    \begin{subfigure}[b]{0.8\textwidth}
        \includegraphics[width=\textwidth]{figures/app-trajs/aircraft/traj_aircraft_1.pdf}
    \end{subfigure}
    \begin{subfigure}[b]{0.8\textwidth}
        \includegraphics[width=\textwidth]{figures/app-trajs/aircraft/traj_aircraft_2.pdf}
    \end{subfigure}
    \begin{subfigure}[b]{0.8\textwidth}
        \includegraphics[width=\textwidth]{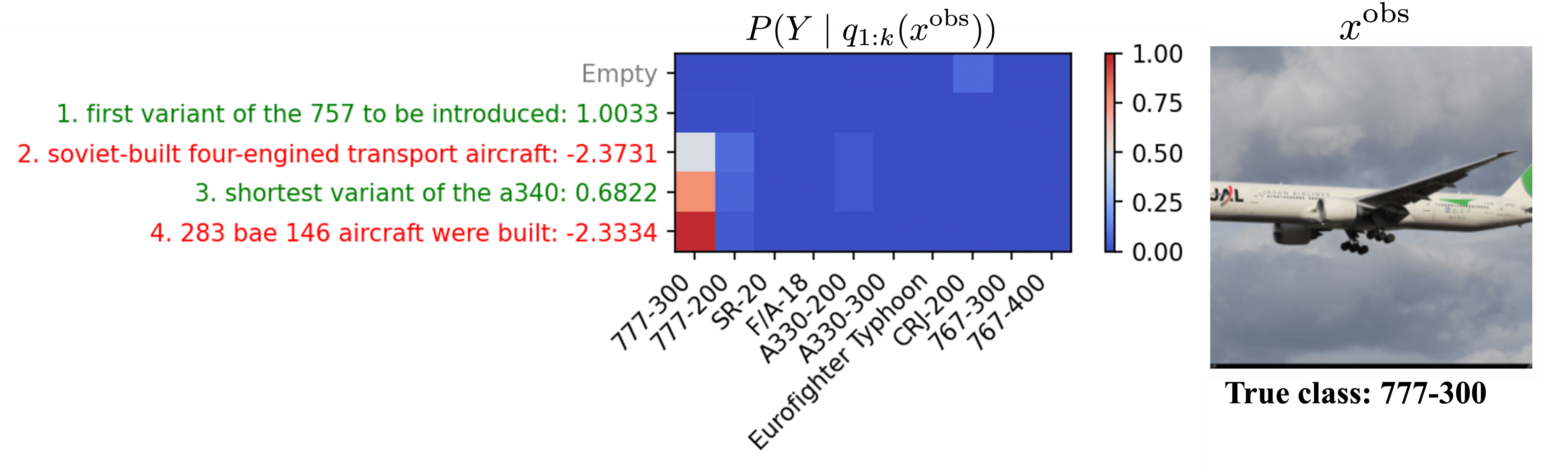}
    \end{subfigure}
    \begin{subfigure}[b]{0.8\textwidth}
    \includegraphics[width=\textwidth]{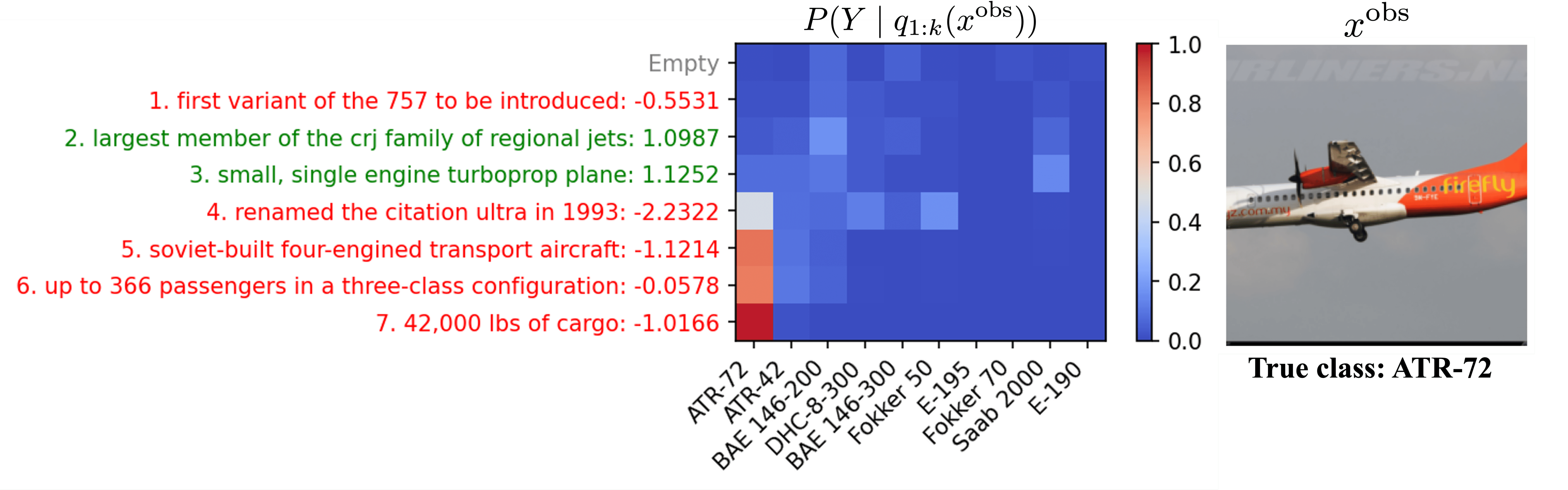}
    \end{subfigure}
    \begin{subfigure}[b]{0.8\textwidth}
    \includegraphics[width=\textwidth]{figures/app-trajs/aircraft/traj_aircraft_5.pdf}
    \end{subfigure}
    \caption{Five examples of query-answer chains from FGVC-Aircraft.}\label{app:traj-aircraft}
\end{figure}

\begin{figure}
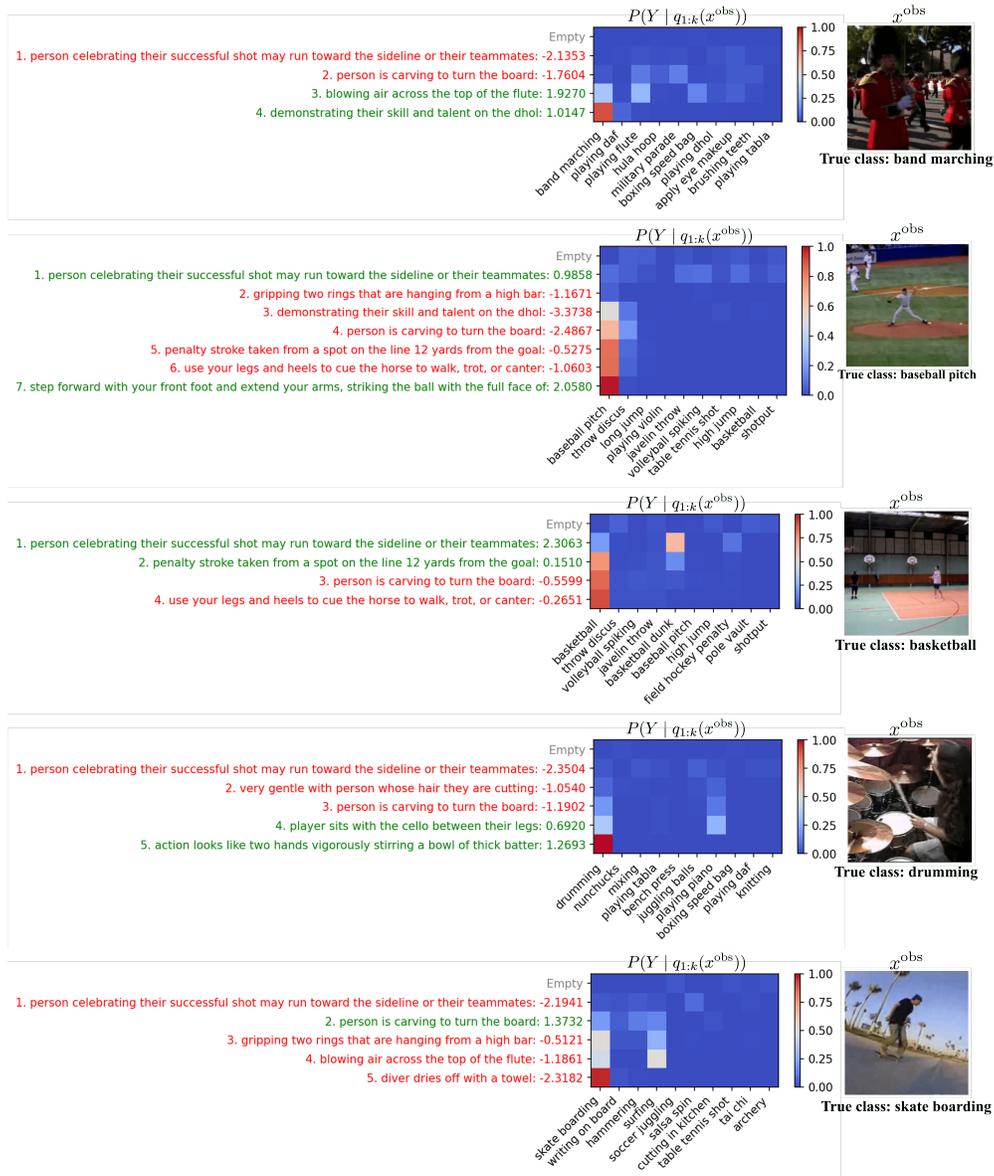

    \centering
    \begin{subfigure}[b]{0.95\textwidth}
        \includegraphics[width=\textwidth]{figures/app-trajs/ucf101/traj_ucf101_1.pdf}
    \end{subfigure}
    \begin{subfigure}[b]{0.95\textwidth}
        \includegraphics[width=\textwidth]{figures/app-trajs/ucf101/traj_ucf101_2.pdf}
    \end{subfigure}
    \begin{subfigure}[b]{0.95\textwidth}
        \includegraphics[width=\textwidth]{figures/app-trajs/ucf101/traj_ucf101_3.pdf}
    \end{subfigure}
    \begin{subfigure}[b]{0.95\textwidth}
    \includegraphics[width=\textwidth]{figures/app-trajs/ucf101/traj_ucf101_4.pdf}
    \end{subfigure}
    \begin{subfigure}[b]{0.95\textwidth}
    \includegraphics[width=\textwidth]{figures/app-trajs/ucf101/traj_ucf101_5.pdf}
    \end{subfigure}
    \caption{Five examples of query-answer chains from UCF-101.}\label{app:traj-ucf101}
\end{figure}

We illustrate more test examples of query-answer chains using FM+V-IP with the $Q_{\text{GPT}}$ query set from Section~\ref{sec:compare-with-lfcbm} on CIFAR-10~\ref{app:traj-cifar10}, CIFAR-100~\ref{app:traj-cifar100}, CUB-200~\ref{app:traj-cub}, ImageNet~\ref{app:traj-imagenet}, Places365~\ref{app:traj-places365}. We also show test examples using FM+V-IP with LaBo's query set from Section~\ref{sec:compare-with-labo} on Flower-102~\ref{app:traj-flower}, FGVC-Aircraft~\ref{app:traj-aircraft}, and UCF-101~\ref{app:traj-ucf101}. Different stopping criteria is used for each dataset: $\epsilon=0.95$ for CIFAR-10, $\epsilon=0.85$ for CIFAR-100, $\epsilon=0.8$ for CUB-200, $\epsilon=0.8$ for ImageNet, $\epsilon=0.8$ for Places365, $\epsilon=0.9$ for Flower-102, $\epsilon=0.9$ for FGVC-Aircraft, and $\epsilon=0.9$ for UCF-101. For datasets with more than 10 classes, we show only the classes with the top 10  probability class where IP terminates. Similar to other examples of query-answer chains, green/red query (y-axis of the plots) represents whether query answer is above/below 0. 

{ The query set used for Flower-102, FGVC-Aircraft and UCF-101 are from LaBo, which is also available on their github. Since their prompts are not visual-attribute specific, some concepts may appear non-visual related.}


\end{document}